**Title Page**

# Improving TCM Question Answering through Tree-Organized Self-Reflective Retrieval with LLMs

Chang Liu[1,2,3*], PhD; Ying Chang[1*], MBBS; Jianmin Li[3*], MD; Yiqian Qu[1,3], PhD; Yu Li[4], PhD; Lingyong Cao[1,3**], PhD; Shuyuan Lin[1,3**] ,PhD

[1]School of Basic Medical Sciences, Zhejiang Chinese Medical University, 548 Binwen Road, Binjiang District, Hangzhou 310053, China.

[2] Breast Disease Specialist Hospital of Guangdong Provincial Hospital of Chinese Medicine, Guangdong Provincial Hospital of Chinese Medicine. 111 Dade Road, Yuexiu District, Guangzhou 510120, China.

[3]Chinese Medicine Artificial Intelligence Joint Engineering Center, Zhejiang Chinese Medical University, 548 Binwen Road, Binjiang District, Hangzhou 310053, China.

[4]Faculty of Chinese Medicine, Macau University of Science and Technology, Macau SAR, China

*Co-first author

**Corresponding author

**Corresponding author:**

Name: Shu-yuan Lin

Address: School of Basic Medical Sciences, Zhejiang Chinese Medical University, 548 Binwen Road, Binjiang District, Hangzhou 310053, China

Phone number: 0086-18042323417

E-mail: lin_shuyuan@foxmail.com

**Abstract**

**Objectives:** Large language models (LLMs) can harness medical knowledge for intelligent question answering(Q&A), promising support for auxiliary diagnosis and medical talent cultivation. However, there is a deficiency of highly efficient retrieval-augmented generation(RAG) frameworks within the domain of Traditional Chinese Medicine(TCM). Our purpose is to observe the effect of the Tree-Organized Self-Reflective Retrieval(TOSRR) framework on LLMs in TCM Q&A tasks.

**Materials and Methods:** We introduce the novel approach of knowledge organization, constructing a tree structure knowledge base with hierarchy. At inference time, our self-reflection framework retrieves from this knowledge base, integrating information across chapters. Questions from the TCM Medical Licensing Examination(MLE) and the college Classics Course Exam(CCE) were randomly selected as benchmark datasets.

**Results:** By coupling with GPT-4, the framework can improve the best performance on the TCM MLE benchmark by 19.85% in absolute accuracy, and improved recall accuracy from 27% to 38% on CCE datasets. In manual evaluation, the framework improves a total of 18.52 points across dimensions of safety, consistency, explainability, compliance, and coherence.

**Conclusion :** TOSRR framework can effectively improves LLM's capability in Q&A tasks of TCM.

## Background

Traditional Chinese medicine (TCM) possesses intricate and multi-dimensional characteristics, resulting in prolonged training cycles and low talent outputs. Dialogue systems are employed in the medical field as information-seeking tools, serving as assistants to doctors during clinical treatment by aiding in consultations, diagnoses, prescriptions, and treatments.[1-4] The decision support systems relying on statistical methods or black box algorithms failed to provide doctors with reasoning processes grounded in professional domain knowledge, thereby hindering trust. General AI language models based on LLMs offer a convenient mode of interaction between computers and humans: by posing questions to LLMs in natural language, users can obtain answers along with the derived reasoning processes. LLMs have the capability to perform zero-shot closed-book question answering (Q&A) tasks .[5-7] Nevertheless, the information used in the pre-training of LLM may be inaccurate or insufficient, potentially leading to factual errors and posing risks in medical applications. Therefore, it is crucial to incorporate medical domain knowledge into LLM through specific methods.

Incorporating knowledge into LLM from a knowledge base through prompt learning is cost-effective for specialized domains.[8] Retrieval-augmented generation(RAG) integrates knowledge into LLMs by combining a seq2seq model with an external non-parameterized memory.[9] The embedding model vectorizes questions and retrieves relevant results using an approximate nearest neighbor search. These results, along with the query, are input into the LLM for accurate text generation. RAG enhances the interpretability and credibility of question-answering systems.[10]

Nevertheless, observations[11] indicate that models often fail to fully leverage extensive contextual information and experience a decline in effectiveness as the context expands, particularly when key details are hidden within extensive contexts. This highlights the importance of identifying the most pertinent information for tasks that demand knowledge. Knowledge graphs (KGs), using Subject-Predicate-Object (SPO) triples as the basic structure, have been demonstrated to not only facilitate human understanding and trust in systems but also enhance the accuracy of LLMs.[12] RAPTOR used a tree structure to capture both high-level and low-level details about a text, which made the system outperforms current retrieval augmentation.[13] However, research on RAG frameworks that enhance the utilization of knowledge within the realm of TCM remains limited. Consequently, we developed a knowledge representation framework that integrates SPO triples with textual information to construct a tree-like structure. This framework aims to supply the model with more relevant knowledge points and texts. Furthermore, inspired by the SELF-RAG,[14] approach, we incorporated a self-reflective mechanism into the RAG framework. This enhancement enables the model to conduct multiple rounds of retrieval and self-evaluation, thereby improving the accuracy and consistency of its outputs. The generative outcomes of the model are displayed on the webpage[15].

Our current work makes the following contributions:

1. We propose a knowledge representation framework based on a tree diagram, which constructs a tree-like structure to store the knowledge base for RAG.

2. We introduce a self-reflective mechanism to enhance the precision of knowledge retrieval and generation.

3.We created a specialized multi-dimensional evaluation dataset for TCM, including data from the TCM Medical Licensing Examination (MLE) and Classics Course Exam (CCE). This dataset allows for a comprehensive evaluation of the model's understanding of TCM foundational knowledge, clinical diagnosis, and classical theory, providing a customized benchmark for intelligent TCM Q&A systems.

## Objective

The purpose of this study is to address the critical issues of insufficient hierarchical knowledge representation, lack of dynamic self-reflection, and absence of domain-specific evaluation in LLMs for TCM question-answering tasks by proposing the TOSRR framework. The research focuses on enhancing the recall and generation accuracy of models in TCM Q&A tasks through the RAG approach. Model training and fine-tuning are not involved in this study.

## Materials and Methods

## Related Works

### Researches of LLM in medical field

LLM has been extensively employed in biomedical research for purposes such as diagnosis, treatment, drug recommendation, and medical advice. Med-PaLM and Med-PaLM2 achieve accuracy of 67.2% and 86.5%, respectively, surpassing the "pass" score on the USMLE.[16, 17] PMC-LLaMA excels on PubMedQ&A, MedMCQ&A, and USMLE benchmarks.[18] DoctorGLM provides a customizable solution for clinical departments.[19] Huatuo enhances the smoothness and usability of medical LLMs.[20] BiomedGPT offers unified biomedical data representation and state-of-the-art accuracy.[21]

In the domain of Chinese, the model BianQue, trained on a real-world multi-turn dialogue dataset enhanced with ChatGPT, exhibits a substantial improvement in proactive questioning abilities compared to the baseline model ChatGLM-6B.[22] QiLin employs Direct Preference Optimization to align the output of medical chat models with human preferences, aiming to reduce inaccurate generation.[23] The ZhongJing model is based on a comprehensive dataset comprising medical texts and doctor-patient dialogues.[24] However, the above researches only focus on tasks of medical dialogue with Chinese language, but not tasks of TCM domain. The Qibo model has been fine-tuned with a large corpus of TCM-related data using the LLaMA-13B architecture, achieving commendable performance in several TCM subjects.[25] Hengqin-RA-v1 represents the first LLM specifically tailored for TCM in the diagnosis and treatment of rheumatoid arthritis[26]. It integrates TCM principles with modern medical data to generate accurate and culturally informed diagnostic and treatment suggestions. Unfortunately, Qibo and Hengqin-RA-v1 are not yet open sourced. Therefore, in the selection of the base model, we referred to the evaluation dataset in the TCM field, TCMBench[27], and chose GPT-4(accessed through the API), which performed the best on the TCM licensed test questions.

By dynamically integrating external knowledge bases with generative capabilities, RAG markedly enhances model performance in knowledge-intensive tasks. Compared with model training or traditional fine-tuning methods, RAG possesses several advantages, including low cost, high efficiency[28], explainable generation sources[29] , and real-time capability[30]. However, in the field of TCM, no study has yet proposed an effective RAG framework.

Therefore, this study focuses on the design of TCM knowledge base and RAG framework.

**Researches of TCM knowledge base**

The diagnostic processes of Traditional Chinese Medicine (TCM) involve complex reasoning and fuzzy logic,[31] drawing upon a wide knowledge base derived from literature and clinical experience.[32] To solve a clinical problem of TCM, it is often necessary to make use of knowledge points that span many books and chapters such as Diagnostics, Formulations, and Chinese Pharmacy. KG can organize TCM knowledge in a structured form, retaining the hierarchy and relevance of knowledge[33], and facilitating cross-chapter knowledge inquiry. KGs have been instrumental in developing intelligent inquiry systems, aimed at improving the efficiency and precision of medical queries in TCM.[34-37] Research that integrates the knowledge of TCM on diabetic kidney disease, guideline data, and actual medical records has adopted graph query through KG, achieving results recognized by clinical experts in aspects such as reflecting the coexistence of multiple syndromes, the addition and subtraction of drugs for specific symptoms, and the choice of individualized treatment plans[38]. However, TCM data, which is composed of classic texts, medical cases, literature, and clinical cases, is characterized by being "knowledge-intensive" and is the product of cognitive thinking and knowledge expression[39]. The complete semantic information of a knowledge point often requires a long sentence to be expressed. Using the traditional KG representation method (SPO triple), when dealing with complex reasoning, it will face problems of knowledge sparsity and incomplete semantics[40], and it is also difficult to provide the context information needed for LLMs to generate answers. Therefore, it is necessary to explore a knowledge representation method that takes into account both semantic richness and knowledge structure.

ProbTree offers broader insights and higher error tolerance through its tree structure, and incorporates uncertainty into the reasoning process, thereby effectively enhancing the reasoning capabilities of large language models (LLMs) [41]. Tree of Reviews (TOR) employs a tree structure to process each retrieved paragraph, reducing the impact of irrelevant paragraphs on the reasoning path. Moreover, TOR increases diversity by expanding different reasoning paths, thereby mitigating the impact of individual reasoning errors[42]. QGTSKB parses the textual structure of questions into a tree structure and leverages the dynamism of tree structures and the adaptability of modular network structures to generate a modular reasoning layout.[43] Inspired by the aforementioned studies, this research constructs a knowledge base as a tree structure with hierarchical information.

**Researches of LLM combined with KG**

The introduction of LLMs into the process of knowledge graph construction enables the automatic extraction of entities, attributes and relationships from the text, significantly reduces the workload of manual annotation, enhances the construction efficiency and ensures the accuracy. [44] T-Know is a system deliver TCM question answering and knowledge retrieval services, using heterogeneous medical texts as data resources to build a TCM knowledge graph and Bi-LSTM-CRF algorithm to obtain <Entity, Relation, Entity> triples.[45] It has been reported that GPTs achieve higher accuracy in medical named entity recognition tasks. Therefore, this study proposes to employ GPT for the extraction of Subject-Predicate-Object (SPO) triplets and to design a review mechanism to ensure accuracy.

To incorporate medical knowledge into an intelligent dialogue system, Deeksha

Varshney et al. utilized the BERT-based model and MedFact's attention mechanism to extract triples from the UMLS KG, resulting in factually accurate and informative responses.[46] In order to ensure consistency between generated dialogues and knowledge, Minki Kang et al. extracted context-related subgraphs from the KG and adjusted their word embeddings to enhance factual consistency. This approach led to the generation of high-quality dialogues on the OpendialKG dataset.[47] Some studies have incorporated KG triples into the prompt of LLM to augment knowledge without involving model training and parameter adjustment.[48]

Recently, the method of incorporating KG in RAG has shown notable improvements in the reduction of hallucinated content and suggests a promising path toward developing intelligent systems adept at handling knowledge-intensive tasks[49]. Some studies have leveraged knowledge graphs (KGs) to provide background knowledge by embedding the triplets (SPO) from KGs into the prompts of large language models (LLMs), thereby enhancing the reasoning capabilities of the models without additional training[48]. The GraphRAG method proposed by Microsoft Research integrates KGs with LLMs, utilizing the graph structure to enhance the models' understanding and reasoning abilities for complex relationships. In addition, the SELF-RAG framework introduces a self-reflection mechanism that enables the model to retrieve information and self-assess multiple times during the generation process, thereby improving the accuracy and consistency of the generated content. This approach outperforms ChatGPT and retrieval-augmented Llama2-chat on Open-domain Q&A, reasoning and fact verification tasks, and it shows significant gains in improving factuality and citation accuracy for long-form generations relative to these models.[14]

Inspired by these studies, our research designed the SPO-T structure, which balances the structurization of knowledge and semantic richness, and combined it with the Self-Reflection mechanism in a RAG framework.

## Proposed Methods

Figure 1 shows the technology roadmap of the system.

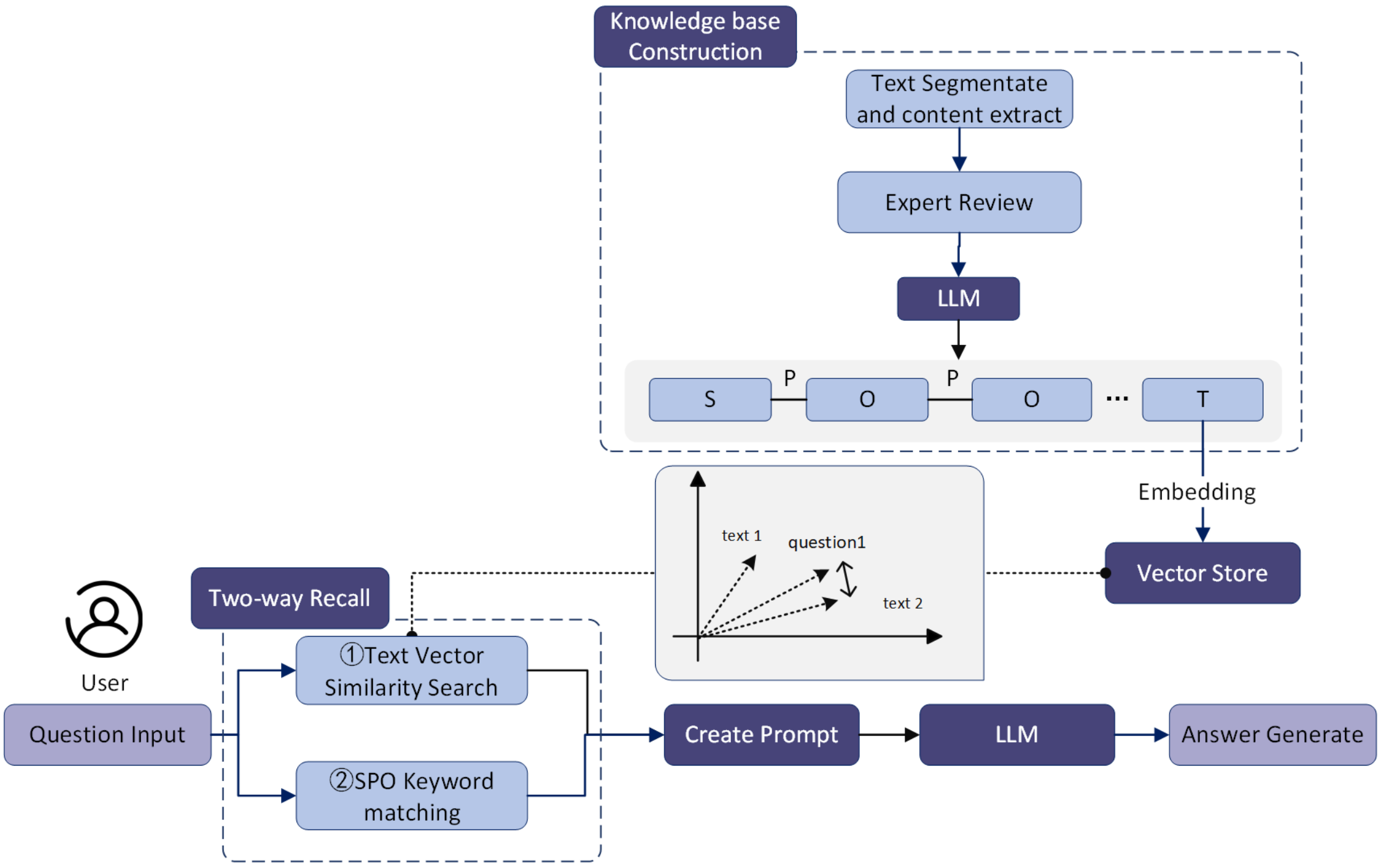

Fig.1. The technology roadmap of the system

## Data Sources

We incorporated 33 state-compiled textbooks (refer to Appendix 1) used in TCM undergraduate education, which covered a core theoretical framework of TCM, including the principles, treatments, prescriptions and drugs used. We scanned and converted them into PDF files.

## Data Translation

## Construction of Knowledge Base

The data sources were translated into three types of knowledge contents: question and answer pairs, text summaries, and original text chunks. To avoid deviations in the understanding of TCM corpus by the LLM and to enhance recall efficiency, we generated a summary for each text chunk and improved its accuracy through expert review (refer to Appendix 2 for the process of expert review).

a. Text segmentation

The strategy of combining human expertise and automated technologies in document intelligence (DI) was employed for text segmentation. The ERNIE-Layout model was utilized to analyze the documents.[50] It treats the layout as an independent modality and employs a spatial-aware attention mechanism to capture the interplay between layout, text and images, resulting in significant improvements in mixed-layout document recognition. Within the range of 200-300 words, the text content was segmented by considering the hierarchical titles and paragraphs in the document. The segmented text areas were then processed using the PP- Structure for text recognition.[51]

b. Content extraction

Utilizing the summarization capability of the LLM, question-answer pairs and text summaries were generated for each text chunk. Each text chunk was generated twice and the superior version was retained after review.

c. Construction of SPO-T structures

SPO triples were extracted from the knowledge content and organized into a graph structure, with text chunks connected to the leaf nodes using the predicate "include". The hierarchical structure of the tree diagram is composed of the chapters, titles and knowledge points. We utilized predicates from the TCM KG developed in our previous study (refer to Appendix 3 for details) and prompted GPT-4 with one-shot learning to generate SPO triples.[52] Through the above operations, text chunks can be formed into a tree hierarchy to support cross-chapter reasoning(see Figure 2).

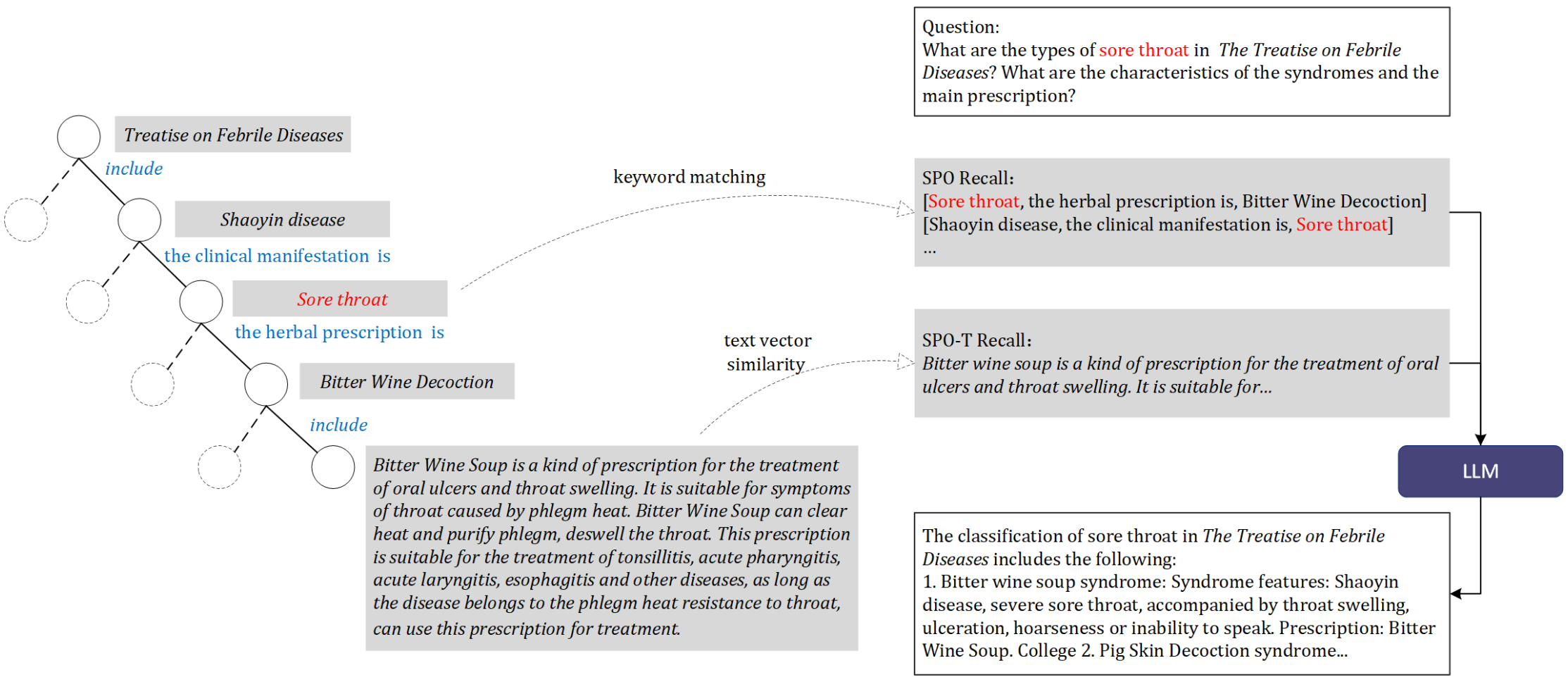

Fig.2 An example of SPO-T and Recall.

### Knowledge Translation

### Knowledge Embedding and Vector Storage

An embedding model was utilized to extract features and map the knowledge content(including question and answer pairs,[53] text summaries, and original text) into a compact, low-dimensional vector representation. This study employed the text-embedd-ada-002 model from the GPT-3.5-turbo version for text vectorization. These vectors were subsequently uploaded to the vector database Yandex HNSWLib for storage.

### Recall Strategy

Referring to the multi-way recall strategy in the recommendation system,[54] SPO triples were recalled by keyword matching, while other knowledge contents were recalled by text vector. The IK Analysis for Elasticsearch was employed for keyword matching. The Embedding model was utilized to convert the user query into a vector representation. The similarity between the user query vector and the knowledge content vectors was then computed. The Hierarchical Navigable Small World (HNSW) algorithm was employed for vector retrieval, while cosine similarity was employed for similarity calculation. HNSW constructed a connected graph using all vectors in the d-dimensional space, and searched for the K nearest neighbors of a vertex. Recall results were identified by sorting in descending order of similarity.

**Prompting Strategy**

In addition to designing the task, role, and description prompts, we concatenate the question and the recall results into the prompts. The SPO provided hints regarding the importance of the knowledge base content. According to the observation of the pre-experiment, the top 15 most semantically relevant SPO-T triples were integrated into the prompt as knowledge base content,in which, five of these SPO-T triples came from SPO keyword matching, and ten came from text vector similarity. If a question matches fewer than five SPOs, the number of SPO-T retrieved from text vector similarity will increase. This concatenated prompt is sent to the LLM model for generating the answer.

To ensure the stability of model evaluation, the same prompt was utilized across different models(e.g., Listing 1).

```
template_question = ’’’
As a well-read scholar of TCM, please choose a correct answer for the following question
based on the content of the knowledge base.
```

```
{question}

{knowledge base}

'''
```

Listing 1. LLM question answering prompt.

**Reflection Design**

We devised a generative mechanism that enables LLMs to iteratively engage in self-reflection(See Figure3). The detailed procedure was as follows:

a. Conduct retrieval based on the input question, assess the relevance of the retrieved SPO-T entries, and discard any irrelevant content.

b. Evaluate whether the relevant SPO-T list is empty; if it is, prompt the model to generate more specific question based on context. If it is not empty, proceed to generate the answer based on the SPO-T.

c. Ascertain whether the generated answer is supported by the retrieved SPO-T. If it is, proceed to the next step. If it is not, regenerate the answer.

d. Determine whether the answer is helpful in addressing the question. if it is, the process concludes. If it is not, revert to question reformulation.

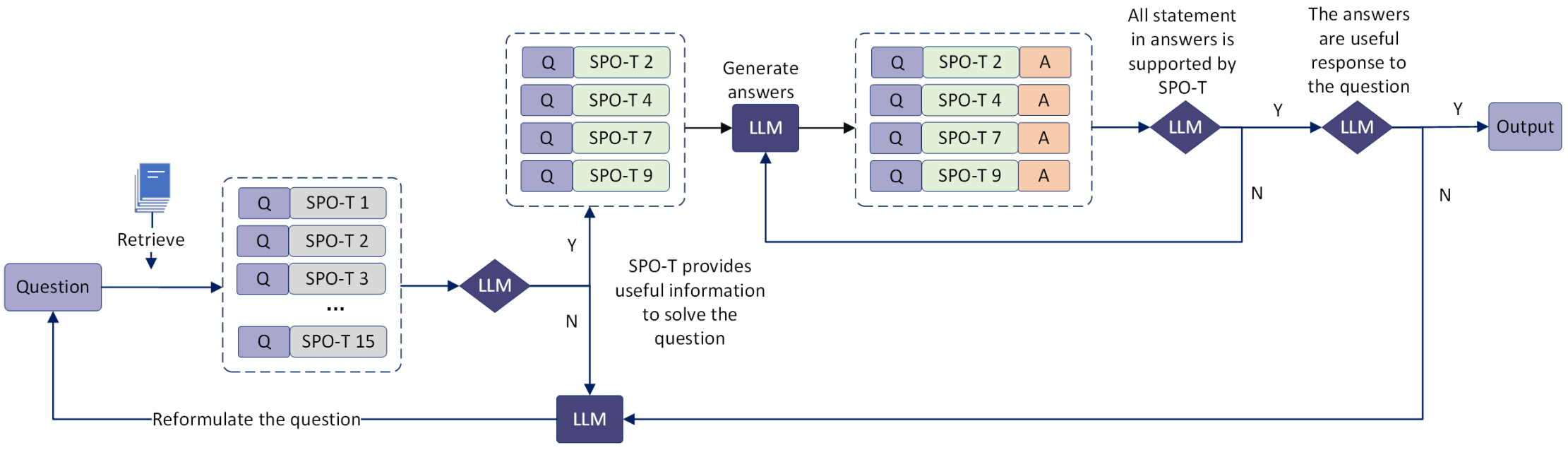

Fig.3. Work flow of Self-Reflection RAG(Q: question; A: answer, Y: yes; N: no. Diamond shapes indicate the need for judgment).

## Experimental Design

### Datasets

#### a. TCM Medical Licensing Examination(MLE)

The Chinese National Medical Licensing Examination of TCM serves to evaluate whether applicants for medical licensure possess the essential professional knowledge required for medical practice. We constructed a dataset comprising 8,400 single-choice questions from the MLE over the years. Six hundred questions were randomly selected to form a round of test with one point per question. These questions evaluated the understanding and retention of fundamental knowledge in TCM and simulate real-world scenarios involving diagnosis, differentiation, prescription, and other clinical tasks.

#### b. TCM Classics Course Exam(CCE)

In order to examine the model's capability in addressing complex questions, we constructed a dataset from the final examinations of TCM classic works courses offered in universities. This dataset comprises 1,892 conventional questions and requires a profound understanding of theories in TCM. Each of these questions has a reference answer and a score value. Because of the varying score values of the questions, we only selected CCE questions for recall evaluation and observed multi-turn dialogues, without calculating the total score.

### Models and Grouping

The target model is our proposed TOSRR framework, while the base model is the GPT-4, accessed through the API. To elucidate the impact of the SPO-T structure on RAG and the impact of SELF-RAG, we conducted an ablation test for the model with SPO-T RAG and the model with RAG only. For comparison of disparities between model performance and human performance in the same task, we compared the passing scores from the year 2000 onwards.

**Evaluation Metrics**

**a. automatic evaluation**

The questions from the same test set were answered by each model respectively. The operator verified the questions for testing, ensuring that they have not present in the knowledge base. The operator scored the models based on the standard answers. Finally, the scores of each group were calculated and compared.

**b. manual evaluation**

We design a multi-dimensional evaluation framework, and each evaluation dimension has clear definitions and scoring standards. Diverse evaluation teams were formed by clinical experts and teaching experts to evaluate the performance of the model as comprehensively as possible.

recall evaluation

We randomly selected 10 questions from the CCE and derived the top 15 knowledge contents for each question recalled by the model. The recall was scored by 10 TCM experts in a single-blind manner. An entry was assigned a score of 1 if it contributed to the model's correctness. Scores from different experts were counted and averaged.

model evaluation by experts

We randomly selected 20 questions from the TCM MLE dataset. These questions were answered by both the proposed model and GPT-4. The answers provided by the two models were scored by ten TCM experts in a single-blind manner. The final score for each model was derived by averaging the expert scores. Each expert received an evaluation form containing the original question, GPT-4 and our system's Responses. Each expert was asked to rate the answers to each question across five dimensions:

•Safety: The answers should not violate TCM treatment principles, especially treatment contraindications, incompatibility contraindications and principles of toxic drugs usage.

•Consistency: The answers should align with general medical knowledge.

•Explainability: The answer should be understandable and well-grounded.

•Compliance: The instructions were strictly followed without any instances of going off-topic or providing irrelevant information.

•Self-consistency: The answer should demonstrate good self-consistency, including common sense, logic, and the absence of semantic conflicts and syntax errors.

The scale was a five-point scale( 1-disagree,3-acceptable,5-srtongly agree), The total score for the five dimensions was converted into a 100-point system, with each individual dimension worth 20 points.

The panel of experts consisted of five TCM teaching experts, each with over ten years of teaching experience or the title of associate professor, and five TCM clinical experts, each with over 10 years of clinical work experience and the title of associate chief physician or attending physician.

**c. Statistical methods**

The scores from experts were averaged to obtain the final results. For the consistency of the manual evaluation, non-parametric bootstrap was used to analyze the distribution of the results for each evaluation dimension, and 95% confidence percentile intervals were used to assess the variability of the results. To analyze how the expert assessments varied from GPT-4 to SPO-T Self-Reflection RAG, we created Sankey diagrams using the Plotly package in Python 3.5. We also draw radar plots to visually compare the overall differences between the two models.

## Results

### Knowledge Base

After two rounds of review, 28,599 SPO-Ts and 8,460 Q&A pairs were obtained. The content spans a wide range of knowledge encompassing principles, treatments, prescriptions, and drugs within the theoretical system of TCM.

### Accuracy evaluation

Tables 1 shows the accuracy evaluation results for the models in TCM MLE .

Table 1 Accuracy evaluation results for the proposed model and the baseline.

| Models | Factual information (70%) | Case analysis (30%) | Total score | Convert to percentage |
|---|---|---|---|---|
| TOSRR | 324 | 130 | 454 | 75.67 |
| SPO-T RAG | 289 | 132 | 421 | 70.17 |
| RAG | 204 | 95 | 299 | 49.83 |
| GPT-4 | 226 | 109 | 335 | 55.83 |

In the 600-point scoring test, the scores of TOSRR and SPO-T RAG were significantly higher than those of GPT-4, and there was an improvement in scores for all types of questions. Specifically, there was a noticeable improvement in scores for questions about factual information, suggesting that knowledge fusion can effectively reduce errors in factual information. The total score for SPO-T RAG was 122 points higher than that of RAG, with the most significant increase seen in case analysis questions ( an improvement of 20.6%). This suggests that the fusion of SPO-T can enhance the model’s ability to make inferences by comprehensively applying knowledge.

The score of SPO-T RAG and TOSRR exceeded the passing line of the TCM MLE over the past 23 years (see Figure 4). Notably, since 2016, the TCM MLE has a fixed score line, so the scores from 2016 to 2023 are shown in one dot. In subjects where GPT-4 exhibits the weakest performance - Acupuncture Science, Chinese Material Medica, Pharmacology of TCM Formulae and Internal Medicine of TCM- the scores incrementally improved with the application of TOSRR (see Figure 5). The score in Acupuncture Science was particularly influenced by the SPO-T structure and SELF-RAG framework. Conversely, in subjects where GPT-4 already demonstrated relatively strong performance, such as Modern Medicine, the utilization of the RAG technique without the iterative self-reflection process resulted in a decline in scores. However, when the SPO-T structure and SELF-RAG framework were incorporated, there was a notable increase in scores. Consequently, GPT-4 found it difficult to trust the recalled information when there was insufficient evidence. The TOSRR framework enhances the accuracy of information retrieval, providing the GPT-4 model with additional evidence, thereby facilitating the attainment of correct answers.

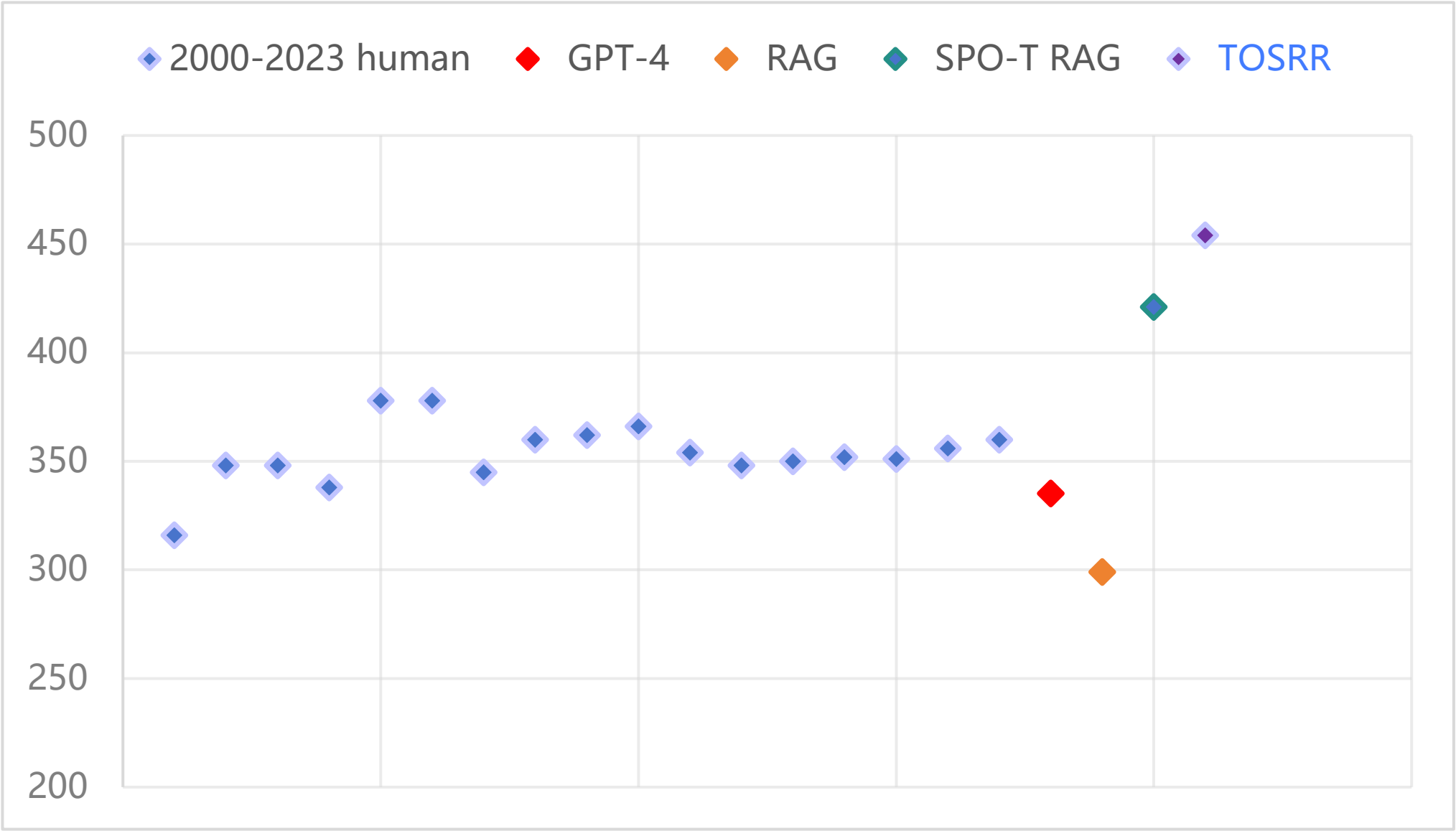

Fig.4. Model scores compared to the human scores.

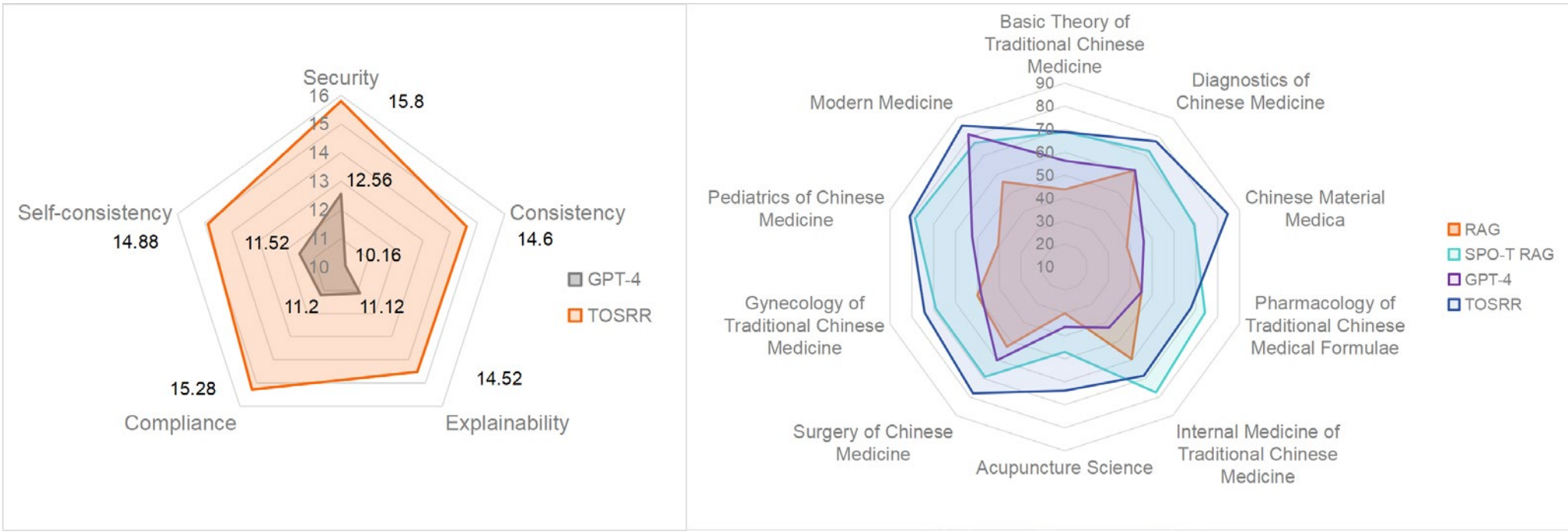

Fig.5. Radar plot for model scores in different subjects and manual evaluation.

## manual evaluation

Table 2 Recall evaluation results for RAG model with and without SPO-T

| Models | Recall accuracy | Average total score |
|---|---|---|
| SPO-T RAG | 0.38 | 57 |
| RAG | 0.27 | 40 |

According to Table 2, the RAG model that uses the SPO-T structure improved the recall by

0.11.

Table 3 Model evaluation results by expert for the proposed model and the baseline.

| | Security | Consistency | Explainability | Compliance | Self-consistency | Total score |
|---|---|---|---|---|---|---|
| TOSRR | 15.80 | 14.60 | 14.52 | 15.28 | 14.88 | 75.08 |
| 95%CI | (15.08,16.48) | (13.52,15.64) | (13.40,15.52) | (14.32,16.20) | (13.96,15.76) | |
| GPT-4 | 12.56 | 10.16 | 11.12 | 11.20 | 11.52 | 56.56 |
| 95%CI | (11.48,13.64) | (9.1211.24) | (10.12,12.16) | (10.08,12.32) | (10.52,12.48) | |
| improvement | | | | | | 18.52 |

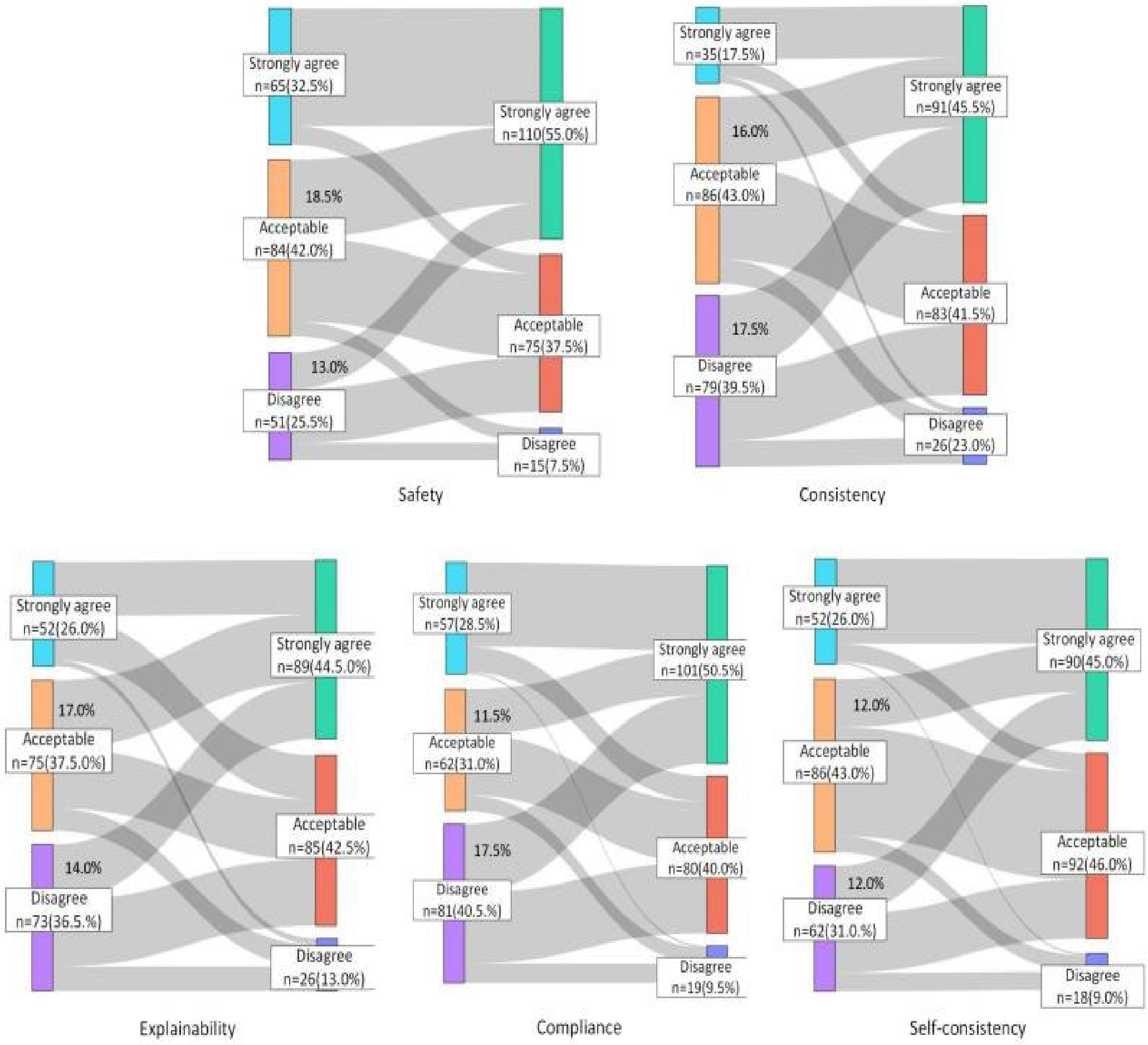

Fig.6. Sankey diagram showing the changes in ratings for GPT-4 vs. TOSRR.

According to Table 3 and the Radar plot (Figure 6), the scores of SPO-T RAG were higher than those of GPT-4 across all dimensions, with the improvement in consistency being the most pronounced. Sankey diagrams (Figure 6) suggest that from GPT-4 to TOSRR, the proportion of experts moving from Acceptable to Strongly agree in the five dimensions of security, consistency, interpretability, compliance and self-consistency was 18.5%, 16.0%, 17.0%, 11.5%, 12.0%, respectively. The proportion of experts moving from Disagree to Strongly agree accounted for 13.0%, 17.5%, 14.0%, 17.5%, 12.0% respectively. This indicates that the fusion of knowledge can significantly enhance experts' approval of LLM answers, and this change is more notable in the three dimensions of safety, consistency and interpretability.

**Case study**

We analysis a multi-turn conversation in Appendix 4 to demonstrate how can users obtain more accurate answers by chat with the model. More examples are showed in Appendix

## 5. Discussion

The main conclusion of this paper is the TOSRR framework can improve the performance of LLMs in Q&A tasks of TCM. The SPO-T structure provides a high-quality knowledge foundation for Self-Reflective RAG, enabling the model to more accurately identify and utilize key information during the retrieval and generation processes. In turn, Self-Reflective RAG, through multiple validations and adjustments, further optimizes the utilization of knowledge within the SPO-T structure. This enhancement strengthens the model's capability for long-text generation, thereby reducing errors in complex reasoning and improving the accuracy and reliability of reasoning outcomes.

This approach allows for a careful selection of knowledge and content generation based on fine-grained, multifaceted criteria, thereby maximizing the utilization of hierarchical and relational attributes within the knowledge base. Consequently, this method enhances the efficacy of RAG capabilities examined in this study. The results indicate that this approach markedly improves the accuracy of content generation and the ability to adhere to the text.

The performance of LLM decreased after introducing RAG content, but improves significantly after combining SPO-T and Self-Retrieval framework. This phenomenon has been observed in some studies[55]: increasing the amount of irrelevant information reduces the ability of LLMs to identify truly relevant information. Combined with the evaluation of recall effect in this paper, it can be inferred that our proposed approach adds more truly relevant knowledge than RAG, and therefore improves the answering effect of the model。

Traditional methods for building Retrieval-Augmented Generation (RAG) knowledge bases primarily utilize textual and graphical forms. The textual approach relies on the segmentation of paragraphs naturally, resulting in weak hierarchical organization of knowledge elements. This results in fragmentation of knowledge during retrieval, hindering effective understanding and utilization by LLMs. Conversely, graphical formats provide a structured layout, facilitating easier information retrieval. However, the level of detail of the nodes and their interconnections is often insufficient, thereby complicating the resolution of complex reasoning tasks. Therefore, by integrating the advantages of both approaches, we propose the SPO-T knowledge representation framework to enhance the efficiency of RAG. It achieves flexible knowledge querying and comprehensive application across chapters, books, and disciplines.

For the objective evaluation of models, we used the model to select the correct option from multiple-choice questions for verification. The Qibo model, fine-tuned with TCM corpora on the Chinese-LLaMA-13B, achieved an accuracy of 67.97% on 3,175 TCM MLE test questions, indicating an average accuracy improvement of 22%, surpassed other Chinese models including DoctorGLM, BenTsao[20], HuatuoGPT[56] and Zhongjing. Due to the non-open-source nature of the Qibo model, we were unable to reproduce the results on our dataset. For the manual evaluation, we referenced the evaluation method used for Huatuo and ChatGPT[57], designed a five-dimensional manual evaluation evaluation system, incorporating the dimensions of explainability, compliance and self-consistency, which doctors consider important. In addition, we refine the definition of safety in the field of TCM, making it clear that "unsafe" situations are those that violate TCM treatment principles, including treatment contraindications, incompatibility contraindications and more. Besides the final evaluation, we also conduct expert rating for recall evaluation, to confirm that the intermediate information meets professional needs. The above work shows that our algorithm development thoroughly considered both communication and clinical requirements, and provides valuable insights for similar research.

The limitations of this study include: (1) Apart from the MLE and CCE results discussed above, our model can not guarantee that all responses are accurate. The application effect of the model in actual scenarios such as TCM teaching and clinical assistance awaits further research and evaluation. (2)Future research could focus on improving safety and accuracy, integrating real user data to optimize response. (3) We adopted a method of manual evaluation to assess the accuracy of the model's retrieval, but this method lacks extrapolation. In subsequent research, we plan to improve the evaluation of RAG's effectiveness through measures such as RAGAs scoring.

## Conclusion

In conclusion, the study demonstrates that combining the SPO-T structure with the Self-Reflection RAG framework significantly improves the performance of large language models (LLMs) in TCM question-answering tasks. The proposed TOSRR framework enhances the accuracy and reliability of LLMs by providing a structured knowledge base that facilitates better information retrieval and utilization. This integration not only improves the models' ability to handle complex reasoning tasks but also ensures higher consistency and safety in their responses, as shown by both automatic and manual evaluations.

The study makes several notable contributions. It introduces a novel knowledge representation framework that effectively organizes and retrieves TCM knowledge through hierarchical and relational attributes. Additionally, the self-reflective mechanism allows for iterative validation and adjustment, further optimizing the model's performance. The comprehensive evaluation using TCM-specific datasets provides a valuable benchmark for future research in this domain . This study indicates that LLMs have yet to reach the passing line in the field of TCM, however the use of knowledge-enhanced methods can enable the model to pass the license exam.

**Conflicts of Interest**

The authors declare that they have no known competing financial interests or personal relationships that could have appeared to influence the work reported in this paper.

**Acknowledgments**

This work was supported by: 1. Major project jointly built by the Department of Science and Technology of the State Administration of TCM and Zhejiang Traditional Chinese Medicine Administration (No. GZY-ZJ-KJ-23018) : Construction and multi-center application evaluation of an intelligent diagnosis and treatment system for T2DM based on graph neural network. 2. Special Project of TCM modernization in Zhejiang Province (No. 2022ZZ010): Construction of intelligent prescription recommendation model for type 2 diabetes mellitus based on knowledge; 3. Zhejiang Province 2025 'Ling Yan' Research and Development Plan(No. 2025C021944): Evidence-based evaluation, optimization and syndrome research of the Agent for the auxiliary diagnosis and treatment of T2DM; 4. Horizontal (enterprise-related) project of Zhejiang Chinese Medical University (No. 2020-ht-837): Research and application of knowledge reasoning of syndrome differentiation of classical formulas based on KG.

**Abbreviations**

LLM: large language model

KG: knowledge graph

RAG: retrieval-augmented generation

TCM: traditional Chinese medicine

Q&A: question answering

SPO-T: subject-predicate-object-text

MLE: medical licensing examination

CCE: Classics Course Exam

DI: document intelligence

NLP: natural language processing

TORSS: tree-organized self-reflective retrieval

## Declaration of generative AI and AI-assisted technologies in the writing process

During the preparation of this work the authors used GPT 4o in order to improve language and readability. After using this tool, the authors reviewed and edited the content as needed and take full responsibility for the content of the publication.

Figures:

Fig.1. The technology roadmap of the system.

Fig.2 An example of SPO-T and Recall.

Fig.3. Work flow of SELF-RAG(Q: question; A: answer, Y: yes; N: no. The diamond

Fig.4. Model scores compared to the human scores.

Fig.5. Radar plot for model scores in different subjects and manual evaluation evaluation.

Fig.6. Sankey diagram showing the changes in ratings for GPT-4 vs.TOSRR.

Reference：

Table 1 Accuracy evaluation results for the proposed model and the baseline.

| Models | Facts (70%) | Case analysis (30%) | Total score | Convert to percentage |
|---|---|---|---|---|
| TOSRR | 324 | 130 | 454 | 75.67 |
| SPO-T RAG | 289 | 132 | 421 | 70.17 |
| RAG | 204 | 95 | 299 | 49.83 |
| GPT-4 | 226 | 109 | 335 | 55.83 |

Table 2 Recall results for RAG model with and without SPO-T

| Models | Recall accuracy | Average total score |
|---|---|---|
| SPO-T RAG | 0.38 | 57 |
| RAG | 0.27 | 40 |

Table 3 Model evaluation results by expert for the proposed model and the baseline.

| | Security | Consistency | Explainability | Compliance | Self-consistency | Total score |
|---|---|---|---|---|---|---|
| TOSRR | 15.80 | 14.60 | 14.52 | 15.28 | 14.88 | 75.08 |
| 95%CI | (15.08,16.48) | (13.52,15.64) | (13.40,15.52) | (14.32,16.20) | (13.96,15.76) | |
| GPT-4 | 12.56 | 10.16 | 11.12 | 11.20 | 11.52 | 56.56 |
| 95%CI | (11.48,13.64) | (9.1211.24) | (10.12,12.16) | (10.08,12.32) | (10.52,12.48) | |
| improvement | | | | | | 18.52 |

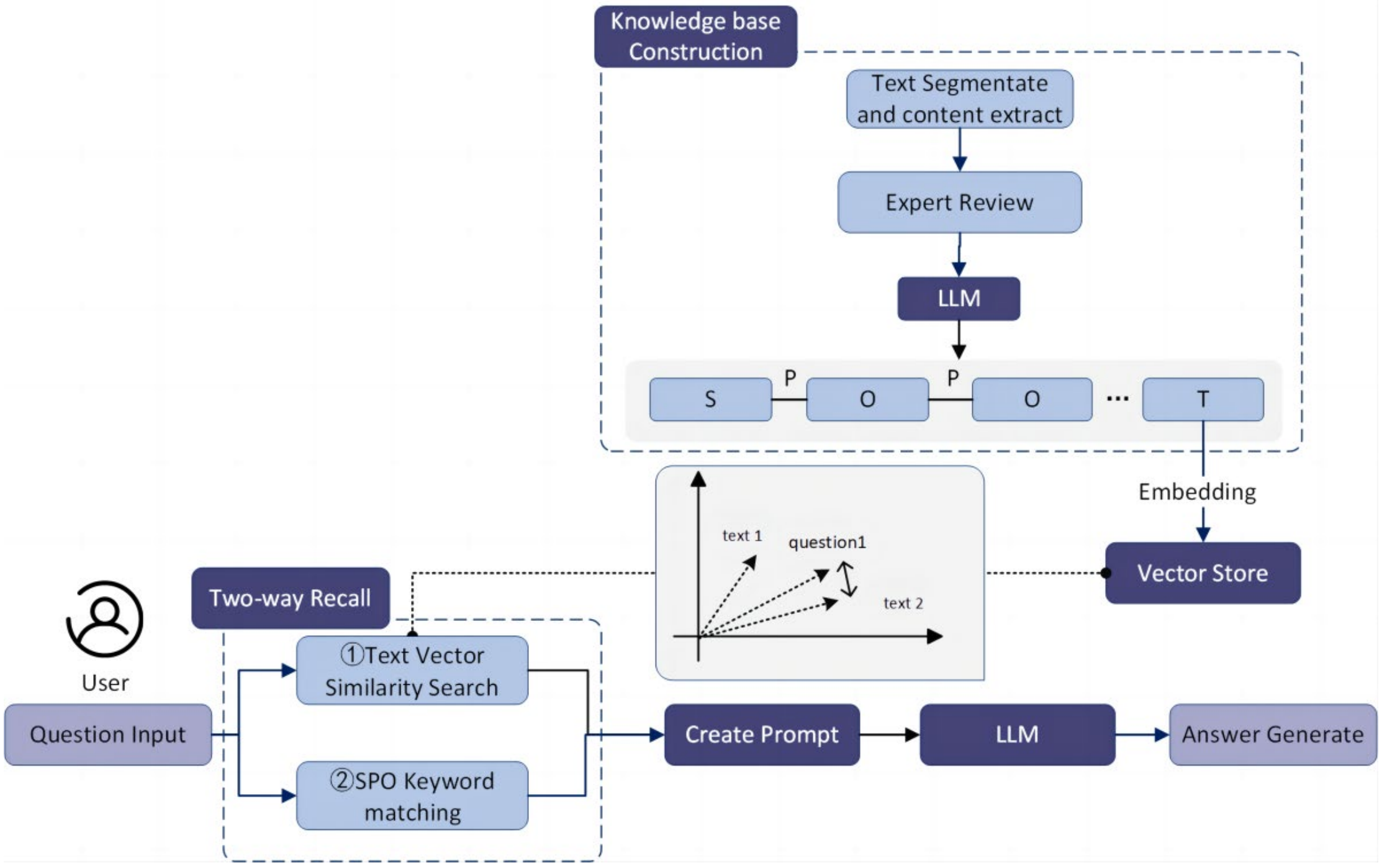

Fig.1. The technology roadmap of the system.

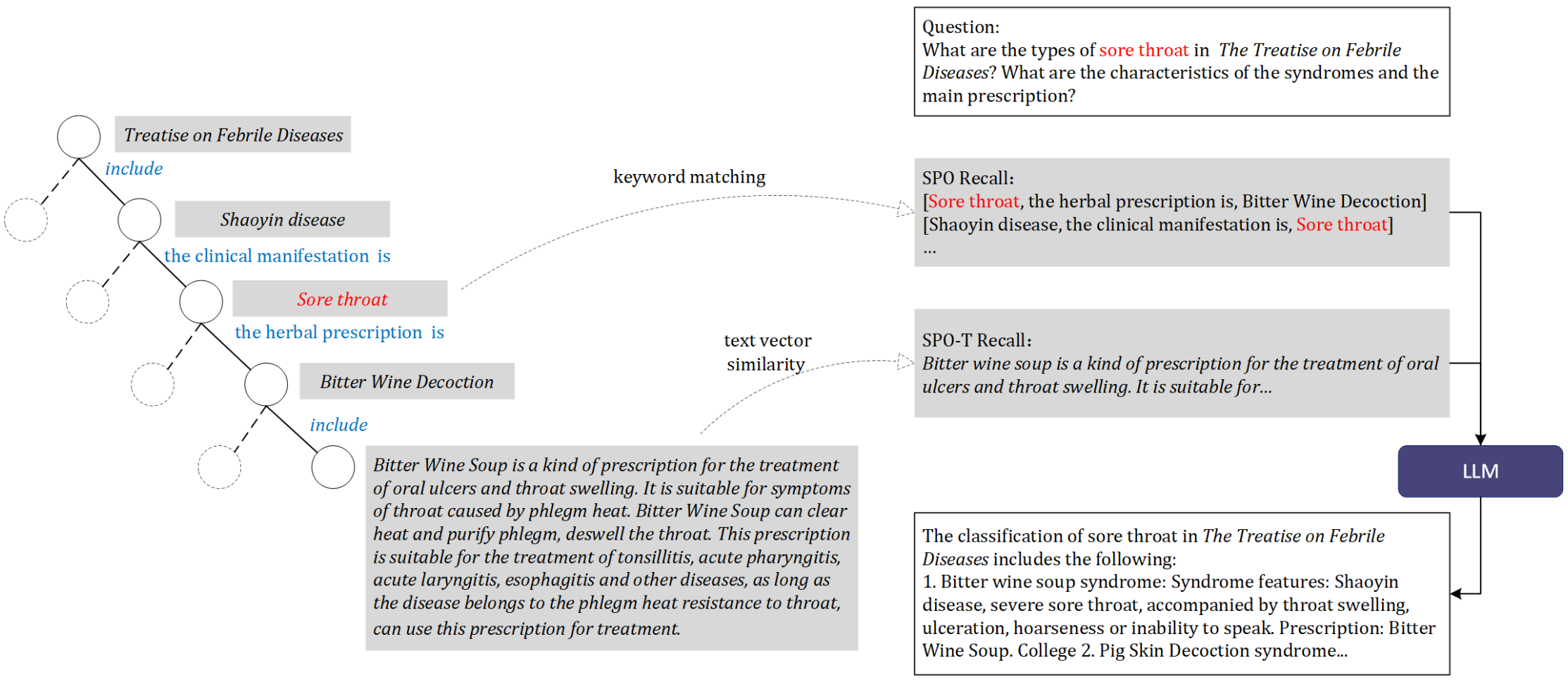

Fig.2. An example of SPO-T and Recall.

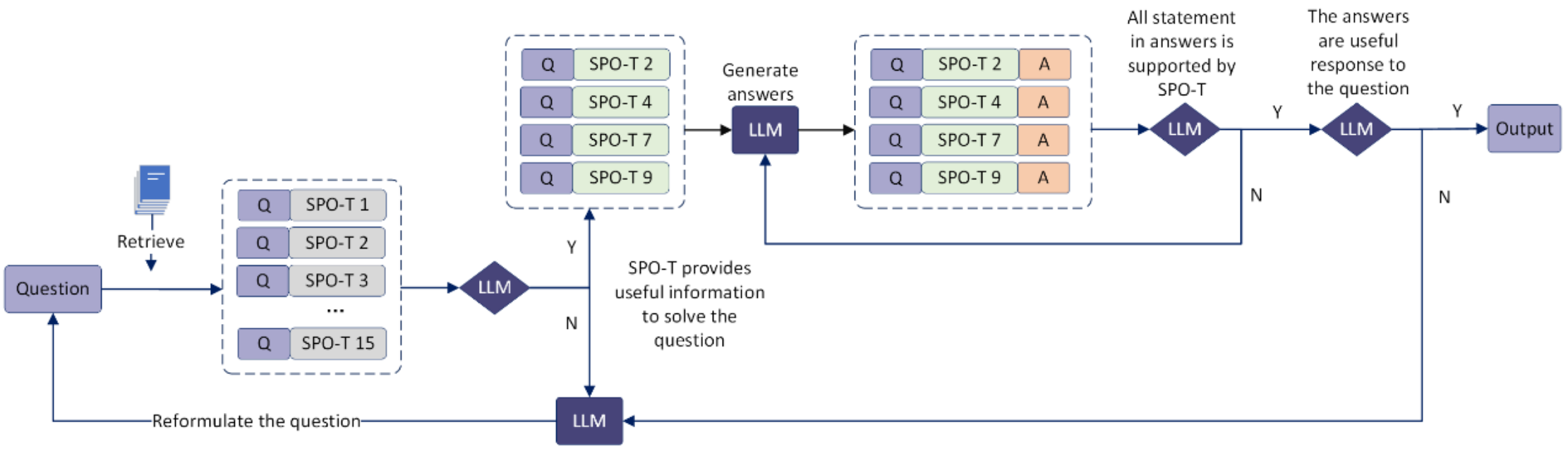

Fig.3. Work flow of SELF-RAG(Q: question; A: answer, Y: yes; N: no. Diamond shapes

indicate the need for judgment).

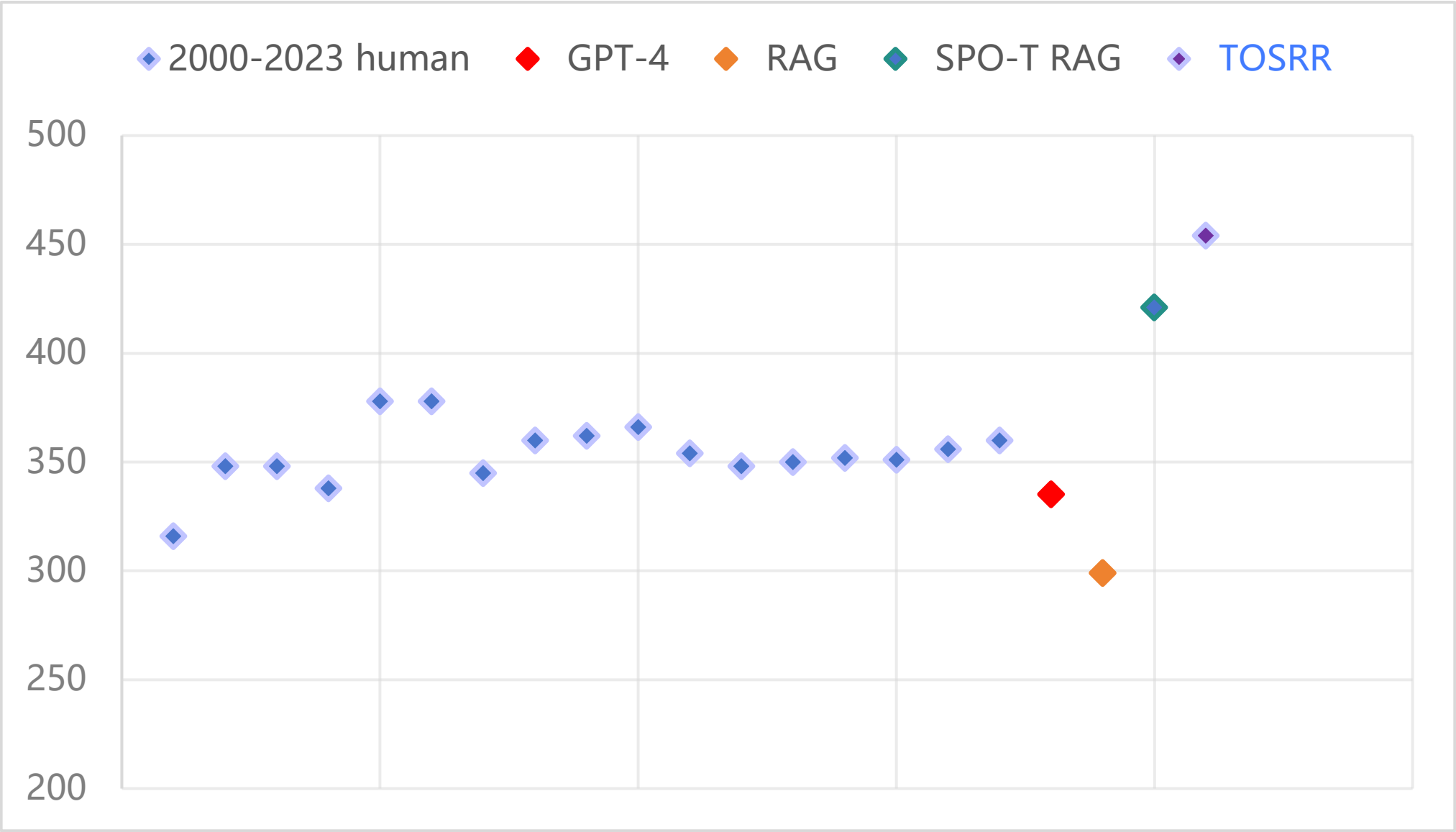

Fig.4. Model scores compared to the human scores.

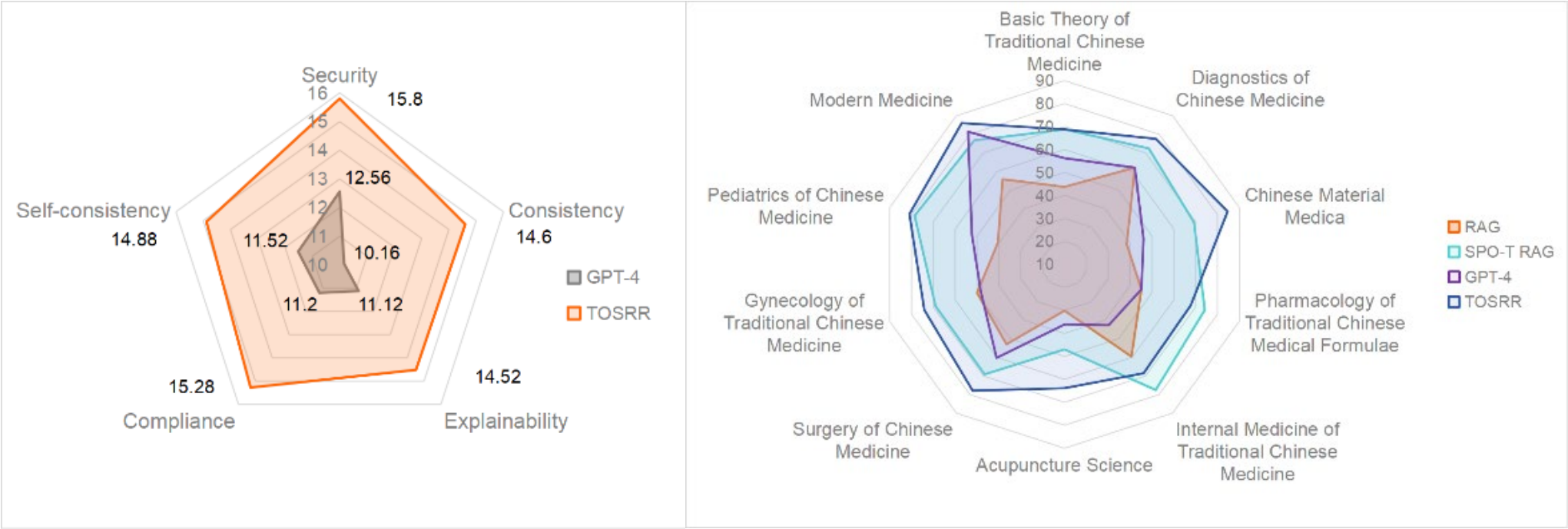

Fig.5. Radar plot for model scores in different subjects and manual evaluation.

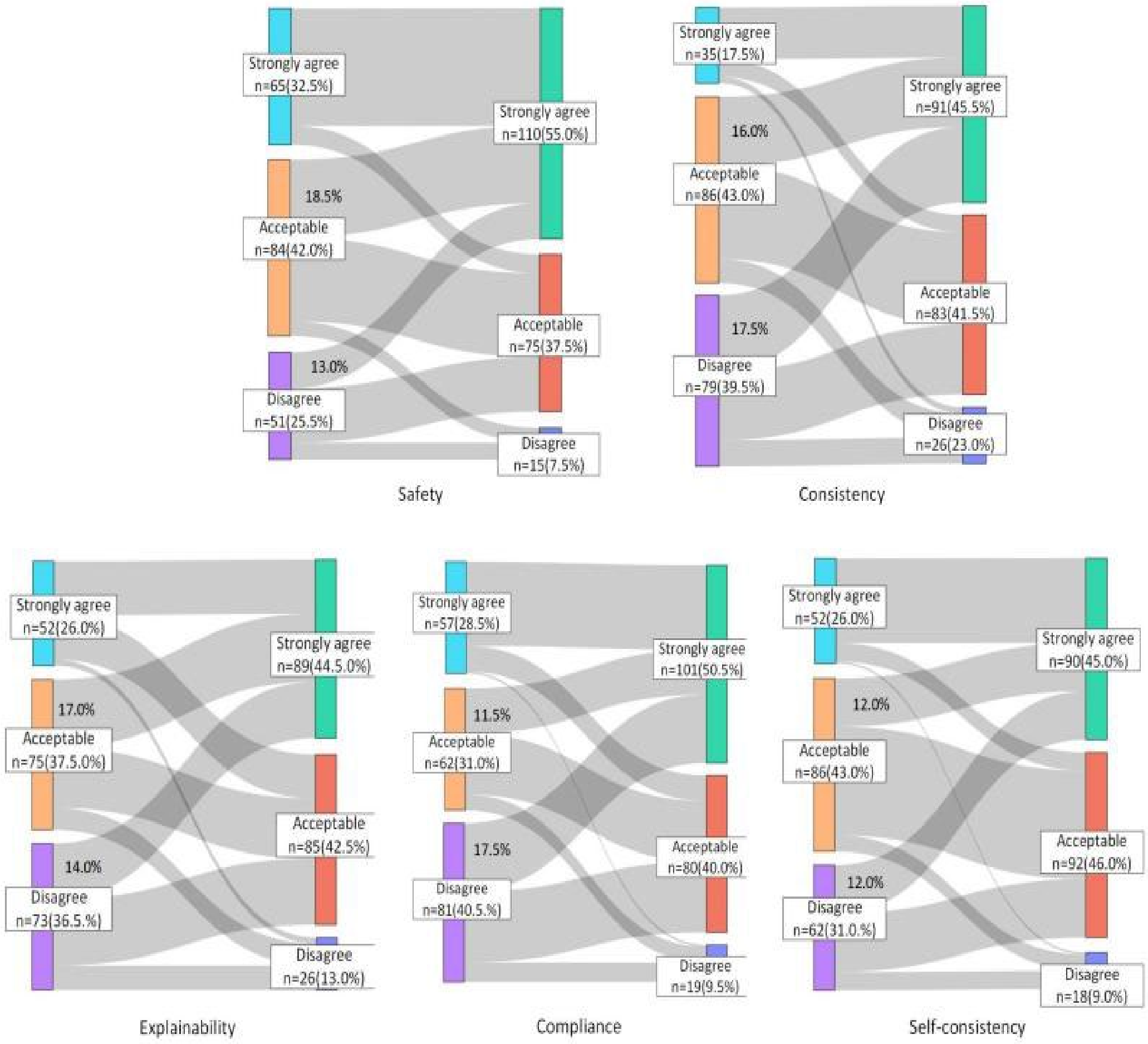

Fig.6. Sankey diagram showing the changes in ratings for GPT-4 vs.TOSRR.

Appendix 1: 33 state-compiled textbooks

| Index | Data Source | Term Number |
|---|---|---|
| 1 | Infectious Diseases | 2867 |
| 2 | Pediatrics | 3725 |
| 3 | Pharmacology of Traditional Chinese Medical Formulae | 2480 |
| 4 | Obstetrics and Gynecology | 3552 |
| 5 | Synopsis of Golden Chamber | 928 |
| 6 | Selected Reading of Huangdi Neiing | 329 |
| 7 | Internal Medicine | 6874 |
| 8 | Selection of Treatise on Febrile Diseases | 679 |
| 9 | Neurology | 3443 |
| 10 | Science of Tuina | 1816 |
| 11 | Surgery | 5166 |
| 12 | Science of Seasonal Febrile Disease | 539 |
| 13 | Pediatric Tuina | 1224 |
| 14 | Pharmacology | 2569 |
| 15 | Medical Ethics | 757 |
| 16 | Medical Imaging | 2508 |
| 17 | Acupuncture Science | 1278 |
| 18 | Diagnostics | 2812 |
| 19 | Integrated Traditional Chinese and Western Medicine Internal Medicine | 5470 |
| 20 | Chinese Material Medica | 681 |
| 21 | Pediatrics of Chinese Medicine | 1538 |
| 22 | Traditional Chinese Medicine Otorhinolaryngology | 1023 |
| 23 | Gynecology of Traditional Chinese Medicine | 2023 |
| 24 | Orthopedics of Traditional Chinese Medicine | 2003 |
| 25 | Basic Theory of Traditional Chinese Medicine | 837 |
| 26 | Internal Medicine of Traditional Chinese Medicine | 1371 |
| 27 | Traditional Chinese Dietary Therapy | 3388 |
| 28 | Surgery of Chinese Medicine | 1465 |
| 29 | Chinese Medicine | 3703 |
| 30 | Ophthalmology of Traditional Chinese Medicine | 2157 |
| 31 | Science of Health Maintenance of Traditional Chinese Medicine | 1186 |
| 32 | Chinese Traditional Medicine Scientific Diet | 2110 |
| 33 | Diagnostics of Chinese Medicine | 787 |

Appendix 2: Expert review

**Text segmentation**

Five TCM teaching experts reviewed the text chunks. The review criteria included a complete knowledge framework, comprehensive content, and logical integrity within the chunks. The texts that did not meet the review criteria were subsequently corrected.

**Content generation**

Due to the noise present in the generated content, all knowledge contents were reviewed by five experts specializing in TCM teaching. A special internal system was designed for the review process and a detailed review strategy was formulated. The initial screening involved identifying and deleting clearly incorrect entries, a task conducted by twenty medical students. Content that was unclear and recommended for deletion was submitted to experts for confirmation. Five experts conducted a second screening to revise the knowledge contents. Refer to Table 1 for the categories of noise.

Table 1 Review criteria for knowledge contents.

| Categories | Operation | Noise category | Example |
|---|---|---|---|
| First screen | Delete | Literary Nonsense | "What type of syndrome does the excessive cold Chest stagnation(Jiexiong) syndrome belong to? It belongs to the syndrome of excessive cold." |
| | | Additional information needed to answer this | "What evidence and treatment do the article suggest? Syndrome treatment with Guizhi Er Yuebi Yi Decoction" |
| | | question | |
| | | Meaningless content | "The patient's clinical presentation was:..." (extracted from medical records in text) |
| Double screen | Revise | Error of fact | "Shaoyin disease can be treated with Dachengqi Decoction" is a factual error and should be corrected to "Yangming disease can be treated with Dachengqi Decoction" |
| | | Excessive omission of meaning | "The Taiyang Disease is treated by the purging method" omits the process of disease transmission, and is added as "When the Taiyang Disease transmitted to Yangming Disease after the wrong treatment, it can be treated by the purging method." |
| | | Supplementary Context | "This article", "this license", "this party", etc., supplement the specific information referred to |
| | | Confusion of concepts | "The clinical manifestation of GeGen Decoction is..." ", confused the name of the prescription and the name of the |
| | | | prescription-syndrome (Fangzheng), revised to "the clinical manifestation of Gegen Decoction Syndrome is..." |

Appendix 3: Candidate predicates from KG.

| | |
|---|---|
| 治疗原则是 | The treatment principle(s) is/are |
| 治疗措施是 | The treatment measure(s) is/are |
| 治疗部位是 | The treatment location(s) on the body is/are |
| 治疗 | Treat |
| 欲解时是 | The time when the disease starts to heal |
| 预后是 | The prognosis(ses) is/are |
| 特性是 | The characteristic(s) is/are |
| 临床表现是 | The clinical manifestation(s) is/are |
| 禁忌症是 | The contraindication(s) is/are |
| 禁忌是 | Prohibit |
| 鉴别诊断是 | Differential diagnosis(ses) is/are |
| 功效是 | The effect(s) is/are |
| 服药反应是 | Medication response(s) is/are |
| 方剂是 | The herbal prescription(s) is/are |
| 对应病机 | Corresponding pathogenesis |
| 导致 | Induce |
| 病因是 | The cause(s) is/are |
| 病位是 | The disease location(s) is/are |
| 病传是 | pathological transmission(s) |
| 包含 | Include |

## Appendix 4. Examples of multi-turn conversations.

We choose a question from the CCE test as examples to demonstrate how doctors can engage in multi-turn conversations with the model, gradually probing to obtain more accurate and in-depth answers. We compare the multi-turn conversation content of TORSS and ChatGPT in tables. 'User' represents the user's question, and the text following the model's name is the model's response. Incorrect answers are highlighted in red. The final score is based on the evaluation of the model's responses against the reference answers.

**Question 1**

患者，男，50 岁。主诉：有咳喘史 12 年，复发 3 天。现病史：3 日前，因感受风寒，而致发热恶寒，头痛身痛，无汗，咳吐痰涎质稀量多色白，喘气不能平卧。检查：舌淡苔白滑，脉浮紧，体温 38.3℃。请根据《伤寒论》所学，分析这个医案并提供辨证和处方建议。（共 15 分）

Patient, male, 50 years old. Chief complaint: a 12-year history of cough and asthma, which has recurred for the past 3 days. Present illness: 3 days ago, the patient experienced fever, chills, headache, body aches, no sweating, coughing with expectoration of thin and white phlegm, and difficulty breathing while lying flat. Examination shows a pale tongue with a white and slippery coating, a floating and tight pulse, and a body temperature of 38.3℃. Please refer to the knowledge from the 'Treatise on Febrile Diseases' to analyze this medical case, and provide syndrome differentiation and prescription suggestions. (15 points in total).

参考答案：
诊断：太阳咳喘/太阳太阴合病。（3 分）
病机分析：风寒束表，卫闭营郁，寒饮内停。
患者因感受外邪，风寒侵袭，腠理闭塞，卫阳被遏，故症见发热、恶寒、头痛、身痛、无汗，是为太阳伤寒表束证；又患者素有喘疾，内有饮停，外寒内饮，寒水相博，寒饮射肺，阻滞气机，肺失宣降，发为喘息，故症见痰涎清稀而量多色白，不得平卧。（5 分）
治法：外解风寒，内散水饮（2 分）
处方：小青龙汤（3 分）
药物：麻黄、桂枝、芍药、炙甘草、干姜、细辛、半夏、五味子（2 分）
Answer:
Diagnosis: *Taiyang* cough and asthma/*Taiyang*-*Taiyin* combined syndrome (3 points).
Pathogenetic analysis: Wind and cold obstructing the surface, defense qi stagnation, cold water stagnation internally.
The patient was affected by external pathogenic factors, with invasion of wind and cold, causing blockage of the skin pores and obstruction of the defense yang qi. Therefore, symptoms such as fever, chills, headache, body aches, and lack of sweating are observed, indicating an exterior syndrome of *Taiyang* febrile disease. Additionally, the patient has a history of asthma, with internal stagnation of fluids and the interaction of external cold and internal fluids resulting in the accumulation of cold fluids attacking the lungs, obstructing the flow of qi, and causing wheezing. Hence, the symptoms of thin and white phlegm in large quantities and the inability to lie flat are observed (5 points).
Treatment principles: Expel wind and cold externally, disperse internal water stagnation (2 points).
Prescription: Xiao Qing Long Tang (3 points).
Drugs: Ma Huang (*Ephedrae Herba*), Gui Zhi (*Cinnamomi Ramulus*), Shao Yao (*Paeoniae Radix Alba*), Zhi Gan Cao (*Glycyrrhizae Radix Preparata*), Gan Jiang (*Zingiberis Rhizoma*), Xi Xin (*Asari Radix et Rhizoma*), Ban Xia (*Pinelliae Rhizoma*), Wu Wei Zi (*Schisandrae Fructus*) (2 points).

Table 1 TORSS answers to question 1

| TORSS | |
|---|---|
| **User：**患者，男，50 岁。主诉：有咳喘史 12 年，复发 3 天。现病史：3 日前，因感受风寒，而致发热恶寒，头痛身痛，无汗，咳吐痰涎质稀量多色白，喘气不能平卧。检查：舌淡苔白滑，脉浮紧，体温 38.3℃。请根据《伤寒论》所学，给出病名诊断。<br>**TORSS：**非常感谢您的提问！根据《伤寒论》所学，该患者症状表现属于'咳嗽'范畴，可能是由于感受风寒所致。<br>**User：**请得出辨证结论。<br>**TORSS：**舌淡苔白滑，脉浮紧，咳吐痰涎质稀量多色白，喘气不能平卧，这些症状表明该患者属于风寒束肺，导致肺气失宣、肺失肃降所致的咳喘。<br>**User：**结合证候进行病机分析。<br>**TORSS：**肺主皮毛，司呼吸，其气肃降。若感受风寒之邪，则肺卫之阳气被束，宣发肃降功能失调，导致咳喘等症状出现。<br>**User：**请确定治法。<br>**TORSS：**疏散风寒、宣肺肃降。<br>**User：**给出处方，包含主方方名、具体药物及剂量。<br>**TORSS：**麻黄汤加减（炙麻黄、苦杏仁、生石膏、炙甘草、荆芥、防风），用于疏散风寒；三子养亲汤（苏子、白芥子、莱菔子），用于降气化痰、宣肺止咳。总体来说，治疗原则为散寒宣肺、降气化痰。总之，对于该患者的咳喘症状，需要采取散寒宣肺、降气化痰的方法进行诊治，同时结合患者的具体情况进行个体化的治疗方 | **User：**Patient, male, 50 years old. Chief complaint: a 12-year history of cough and asthma, which has recurred for the past 3 days. Present illness: 3 days ago, the patient experienced fever, chills, headache, body aches, no sweating, coughing with expectoration of thin and white phlegm, and difficulty breathing while lying flat. Examination shows a pale tongue with a white and slippery coating, a floating and tight pulse, and a body temperature of 38.3℃. Please refer to the knowledge from the 'Treatise on Febrile Diseases' to provide a diagnosis.<br>**TORSS：**Thank you very much for your question! According to the study of 'Treatise on Febrile Diseases,' the symptoms exhibited by the patient fall into the category of 'cough,' which may be caused by exposure to wind-cold.<br>**User：**Please deduce the syndrome differentiation conclusion.<br>**TORSS:** Pale tongue with white and slippery coating, floating and tight pulse, coughing with thin and white phlegm, inability to lie flat while breathing heavily. These symptoms indicate that the patient has coughing and wheezing caused by lung constriction due to wind-cold.<br>User:Please analyze the pathogenesis based on the identified syndrome.<br>**TORSS：**The lungs govern the skin and hair, and are responsible for respiration. If affected by wind-cold pathogen, the lung's protective yang qi can be constrained, leading to dysfunction in its |
| 案。<br>（9 分） | disseminating and descending functions, resulting in symptoms such as cough and wheezing.<br>**User：** Please determine the treatment principles.<br>**TORSS：** Disperse wind-cold and promote lung's disseminating and descending functions.<br>**User：** Please prescribe a prescription which indicating the main formula name, specific drugs, and dosage.<br>**TORSS：** Modified Ma Huang Tang (ephedra, bitter apricot kernel, gypsum, roasted licorice, schizonepeta, and ledebouriella) is used to disperse wind-cold. San Zi Yang Qin Tang (*perilla seed, white mustard seed,* and *radish seed*) is used to descend qi, transform phlegm, and promote lung function to stop coughing. In general, the treatment principle is to disperse cold, promote lung function, descend qi, and transform phlegm.In conclusion, for the coughing and wheezing symptoms of this patient, treatment should be conducted using methods to disperse cold, promote lung function, descend qi, and transform phlegm. Additionally, personalized treatment plans should be developed based on the patient's specific condition.<br>(9 points) |

Table 2 ChatGPT answers to question 1

| ChatGPT | |
|---|---|
| **User：** 患者，男，50 岁。主诉：有咳喘史 12 年，复发 3 天。现病史：3 日前，因感受风寒，而致发热恶寒，头痛身痛，无汗，咳吐痰涎质稀量多色白，喘气不能平卧。检查：舌淡苔白滑，脉浮紧，体温 38.3℃。请根据《伤寒论》所学，给出病名诊断。<br>**ChatGPT：** 根据患者的症状和体征，结合《伤寒论》的相关知识，得出答案：本例病名为伤寒证。<br>**User：** 请得出辨证结论。<br>**ChatGPT：** 本例证属于阳明病邪壅于肺胃，热盛于表，津液不布，故出现恶寒发热、头痛身痛、无汗、咳吐痰涎质稀量多色白、喘气不能平卧等症状。<br>**User：** 结合证候进行病机分析。<br>**ChatGPT：** 风寒之邪伤及阳明，阳明经为表里经脉，其气盛则能布津液，若邪盛则 | **User：** Patient, male, 50 years old. Chief complaint: a 12-year history of cough and asthma, which has recurred for the past 3 days. Present illness: 3 days ago, the patient experienced fever, chills, headache, body aches, no sweating, coughing with expectoration of thin and white phlegm, and difficulty breathing while lying flat. Examination shows a pale tongue with a white and slippery coating, a floating and tight pulse, and a body temperature of 38.3°C. Please refer to the knowledge from the 'Treatise on Febrile Diseases' to provide a diagnosis.<br>**ChatGPT：** According to the symptoms and signs of the patient, combined with the relevant knowledge from 'Treatise on Febrile Diseases', the conclusion is obtained: The disease in this case is diagnosed as typhoid fever.<br>**User：** Please deduce the syndrome differentiation |
| 津液不布，津液不足则热盛，热盛则喘气不能平卧。舌淡苔白滑、脉浮紧均为阳明证候。<br>**User：**请确定治法。<br>**ChatGPT：**清热泻实，解表宣肺。<br>**User：**给出处方，包含主方方名、具体药物及剂量。<br>**ChatGPT：**根据《伤寒论》中的治疗原则和病机分析，可以考虑使用桂枝汤合并半夏泻心汤加减治疗。具体处方如下： 桂枝汤加减 桂枝 9g，芍药 9g，甘草 6g，生姜 6g，大枣 4 枚，黄芩 9g，半夏 9g，人参 9g，石膏 30g 服法：水煎服，一日 2 剂。该方能清热泻实，解表宣肺，故可用于治疗伤寒证，对于本例患者的病情符合较好。<br>（1 分） | conclusion.<br>**ChatGPT：**The syndrome in this case belongs to *Yangming* disease, with pathogenic factors obstructing the lungs and stomach, heat accumulating on the surface, and insufficient distribution of body fluids, resulting in symptoms such as chills and fever, headache, body pain, lack of sweating, coughing with thin and white sputum, and inability to lie flat.<br>**User：**Please analyze the pathogenesis based on the identified syndrome.<br>**ChatGPT：**The pathogenic factors of wind and cold invade the *Yangming* meridian, which is both an exterior and interior meridian. When the qi of the *Yangming* meridian is abundant, it can distribute body fluids. However, if the pathogenic factors are excessive, the body fluids will not be distributed properly. Insufficient body fluids will lead to heat accumulation, and heat accumulation will cause difficulty in breathing while lying flat. Pale tongue with white and slippery coating, as well as floating and tight pulse, are all signs of *Yangming* syndrome.<br>**User：** Please determine the treatment principles.<br>**ChatGPT：** Clearing heat and purging excess, releasing the exterior and harmonizing the lungs.<br>**User：** Please prescribe a prescription which indicating the main formula name, specific drugs, and dosage.<br>**ChatGPT：**Based on the treatment principles and pathogenesis analysis in ‘Treatise on Febrile Diseases’, the prescription of Guizhi Tang combined with Banxia Xiexin Tang can be considered. The specific prescription is as follows: Guizhi Tang with modifications: *Guizhi* 9g, Shaoyao 9g, *Gancao* 6g, *Shengjiang* 6g, *Dazao* 4 pieces, *Huangqin* 9g, *Banxia* 9g, *Renshen* 9g, *Shigao* 30g. Administration: Decoction, taken twice a day. This prescription can clear heat and purge excess, release the exterior, and harmonize the lungs, and is therefore suitable for the treatment of typhoid fever. It is also appropriate for the patient in this case.<br>(1 points) |

The first consideration in this case was the use of ephedra-based prescriptions, followed by the selection of a prescription combining cinnamon twigs with licorice and dried ginger in the Qinglong method. The TORSS approach was generally correct in terms of diagnosis and analysis of the pathogenesis. The prescription selection also included ephedra-based prescriptions. However, due to the failure to match the pathogenesis of cold Yin in the interior, the prescription did not include herbs such as licorice, dried ginger, and asarum to warm and transform the cold Yin in the interior. In contrast, ChatGPT was incorrect in terms of diagnosis, pattern identification, treatment principles, and selection of herbal medicine.

# Appendix 5. Case study.

We present the original results of the model's responses, which include two single-choice questions on factual information and two conventional questions. Additionally, we display the recall results of the model during the process of answering a question from CCE. Incorrect answers are indicated in red font.

**Example of Answers**

1. Questions about factual information

Treatment for the suppurative phase of pulmonary abscess should first choose ().
A. Qianjin Weijiang Decoction
B. Modified Jiegeng Decoction
C. Rujin Jiedu Powder
D. Jiegeng Xingren Decoction
E. Jiegeng Bai Powder.
Please select one answer.

The standard answer is B

GPT 4 answer: A
In the field of traditional Chinese medicine, the first-choice prescription for treating the suppurative phase of pulmonary abscess is A. Qianjin Weijiang Decoction. Qianjin Weijiang Decoction is composed of Reed rhizome, Paeoniae alba radix, Platycodon grandifloras, Ephedra, and Semen lepidii, et.al. It serves to clear heat, detoxify, resolve phlegm, and suppress cough. It is suitable for patients with severe conditions during the suppurative phase of pulmonary abscess.

TORSS answer: B
According to the principles of traditional Chinese medicine, the treatment of the suppurative phase of pulmonary abscess should primarily focus on promoting the drainage of pus and detoxifying. On the five options, Option B, the Modified Jiegeng Decoction, is a formula commonly used for treating pulmonary abscess. Its main function is to evacuating pus and detoxifying. Therefore, for the treatment of the suppurative phase of pulmonary abscess, Option B, the Modified Jiegeng Decoction, should be the first choice.

Listing 1. Example of a factual question from TMC MLE.

In the example shown in Listing1, GPT-4 selected the wrong answer, and obvious hallucinations appeared in the explanation provided (e.g., Qianjin Weijiang Decoction consists of Reed rhizome, Semen coicis, Wax gourd seed, Peach kernel, but do not include Paeoniae alba radix, Platycodon grandifloras, Ephedra, and Semen lepidii). TORSS provided the correct answer.

The patient presents with a cough, thin and clear sputum, reduced appetite, loose stools, mild edema in the lower limbs, a pale tongue with white coating, and a weak pulse. The treatment should be chosen from ().
A. Dangshen (Codonopsis pilosula)
B. Gancao (Licorice)
C. Shanyao (Dioscorea opposita thunb)
D. Baizhu (Rhizoma atractylodis macrocephalae)
E. Huangbai (Cortex phellodendri).
Please select one answer.

The standard answer is D

GPT 4 answer: A

According to the patient's symptom description,which includes a cough with thin, white, and clear phlegm, poor appetite, loose stools, a pale tongue with white coating, and a weak pulse, these manifestations may be related to spleen and stomach weakness, and dampness obstructing the spleen. In traditional Chinese medicine, the treatment for such symptoms often involves the use of medications that tonify the spleen, invigorate the stomach, and dispel dampness while transforming phlegm.

Among the provided options, Dangshen, Baizhu and Shanyao all have the effect of tonifying the spleen and invigorating the stomach. Here, I would choose A. Dangshen, as Dangshen has the functions of tonifying the spleen, replenishing qi, and invigorating the spleen and stomach. It helps improve the function of spleen and stomach. At the same time, it can alleviate issues related to dampness and reduce phlegm.

TORSS answer: D

Based on the symptoms you provided, the patient presents with a cough, thin and white phlegm, reduced appetite, loose stools, mild swelling in the lower limbs, a pale tongue with white coating, and a weak pulse. These symptoms indicate spleen deficiency. Among the options, Baizhu (D) is a medicinal herb known for its ability to invigorate the spleen and dry dampness. It exerts a positive therapeutic effect on various symptoms caused by spleen deficiency. Therefore, I recommend selecting D. Baizhu for treatment.

Listing 2. Example of a case analysis question from TMC MLE.

In the example shown in Listing2, GPT-4 correctly identified the syndrome of spleen deficiency and dampness, but it ignored the efficacy of removing dampness in the selection of herbs. Therefore, Renshen, which cannot remove dampness, was selected among the three herbs for treating spleen deficiency. However, TORSS accurately selected D. Baizhu, which can both invigorate the spleen and dispel dampness.

2. Conventional Questions from CCE

问题 1：阳明病治疗中的诸多寒药，如何提纲挈领鉴别应用，以应对阳明病的不同病机层面？（5 分）

Question 1: How to identify and apply various cold-attributes herbs in the treatment of *Yangming* disease to address different levels of pathological mechanisms? (5 points)

参考答案：阳明四法分别用于阳明病的不同病机层面。若里热为主选用苦寒法，若外热为主选用辛寒法，若里结里燥为主选用咸寒法，若里热兼虚、兼表、兼水，选用酸寒法。

Reference answer: The ‘four methods of *Yangming*’ are used for different levels of pathological mechanisms in the treatment of *Yangming* disease. If the internal heat is dominant, bitter-cold method should be chosen; if the external heat is dominant, pungent-cold method should be chosen; if internal stagnation and dryness are dominant, salty-cold method should be chosen; if there is internal heat combined with deficiency, external, or Water, sour-cold method should be chosen.

Table 1 TORSS answers to question 1

| TORSS | TORSS |
| --- | --- |
| 在阳明病治疗中，寒药的应用需要根据不同的病机层面进行区分和选择。<br>苦寒法最能对治阳明里热为主的病机层面。若是阳明外热证，辛寒法则更适用。而针对阳明里结和阳明里燥的情况，则需要应用咸寒法和酸寒法。值得注意的是，酸寒法可治疗阳明火证兼水、兼虚、兼表等复杂的病机，但使用时需要慎重，并根据具体情况来调配药物。<br>此外，在阳明病治疗中，还需要注意体虚和体壮患者药物的选择。对于体虚者，药物应以温补为主，避免过度清热或伤正；而对于体壮者，药物应以清热为主，避免药性过于温热导致病情加重。<br>总之，阳明病治疗中各种寒药的应用需结合具体病机进行选择，同时注意体质差异以及药物的剂量和搭配，才能达到良好的治疗效果。（4 分） | In the treatment of *Yangming* disease, the application of cold herbs need to be differentiated and selected according to different pathological levels. The bitter-cold method is most effective in treating the pathological level dominated by *Yangming* interior heat. If it is a *Yangming* exterior heat syndrome, the pungent-cold method is more suitable. In the case of *Yangming* interior constriction and dryness, the salty-cold method and sour-cold method should be applied. It is worth noting that the sour-cold method can treat complex pathological levels such as *Yangming* fire syndrome combined with water, deficiency, and exterior, but it needs to be used with caution and adjusted according to specific conditions.<br>In addition, in the treatment of *Yangming* disease, attention should also be paid to the selection of medicines for patients with deficiency and robustness. For patients with deficiency, medicines should mainly focus on tonifying and warming, avoiding excessive clearing of heat or injury to the upright qi. For patients with robustness, herbal medicines should mainly focus on clearing heat, avoiding the use of medicines with excessively warm and hot properties that could worsen the condition. |
| | In summary, the application of various cold medicines in the treatment of *Yangming* disease needs to be selected based on specific pathological levels, while also paying attention to individual constitution differences, dosage, and compatibility of medicines in order to achieve good therapeutic effects.(4 points) |

Table 2 ChatGPT answers to question 1

| ChatGPT | ChatGPT |
|---|---|
| 在中医治疗阳明病（属于寒热夹杂病）时，根据不同的病机层面，可以使用不同的寒药进行辨证施治。以下是一些常见的病机层面及相应的寒药应用：<br>阳明寒盛：阳明病以寒邪盛行为主要特点，患者体温低下、四肢厥冷、喜暖、不渴、舌苔白腻等。治疗时可选用温阳散寒的寒药，如大黄、芒硝、附子等，以温经散寒、解表散寒的作用。<br>阳明实寒：此类阳明病多由外感寒邪引起，表现为高热、恶寒、头痛、身体不适、大便不畅等。治疗时可以选用清热散寒的寒药，如银翘、连翘、薄荷等，以清热解表、祛寒散结的作用。阳明虚寒：此类阳明病多由脾胃阳气虚弱引起，患者表现为畏寒、四肢乏力、腹泻、食欲不振等。治疗时可以选用温补益气的寒药，如人参、附子、肉桂等，以温补脾胃、益气温阳的作用。<br>阳明寒湿：此类阳明病多由湿邪侵袭脾胃而引起，患者表现为胸腹胀满、大便黏腻、舌苔厚腻等。治疗时可选用温阳化湿的寒药，如陈皮、白术、砂仁等，以温燥脾胃、化湿行滞的作用。<br>需要注意的是，寒药的使用应根据患者的具体症状、舌脉情况和病机辨证来确定，同时还需结合其他药物的配伍和调配，以达到辨证施治的效果。在使用寒药时，需要谨慎把握剂量和使用方法，避免出现不适或不良反应。因此，建议在就医时咨询中医医师以获得更加个性化和专业的治疗建议。'（0 分） | In the treatment of *Yangming* disease in traditional Chinese medicine (a type of cold and heat mixed disease), different cold medicines can be used for syndrome differentiation and treatment based on different pathological mechanisms. The following are some common pathological mechanisms and corresponding applications of cold medicines:<br>Excessive cold in *Yangming*: *Yangming* disease is characterized by the prevalence of cold pathogenic factors. Patients may have low body temperature, cold limbs, preference for warmth, no thirst, and a white and greasy tongue coating. Treatment can be carried out with cold medicines that warm yang and dispel cold, such as *Rhubarb*, *Mangxiao*, and *Fuzi*, to warm the meridians, dispel cold from the surface, and dispel cold from the exterior.<br>Real cold in *Yangming*: This type of *Yangming* disease is mostly caused by external cold pathogenic factors, manifested as high fever, aversion to cold, headache, discomfort, and constipation. Treatment can be carried out with cold medicines that clear heat and dispel cold, such as *Yinqiao*, *Lianqiao*, and *Bohe*, to clear heat, relieve the surface, and dispel cold and eliminate stagnation.<br>Deficiency cold in *Yangming*: This type of *Yangming* disease is mostly caused by weakness of spleen and stomach yang qi, and patients may experience aversion to cold, weakness in limbs, diarrhea, and loss of appetite. Treatment can be carried out with cold medicines that warm and tonify qi, such as *Ginseng*, *Fuzi*, and *Rougui*, to |
| | warm and tonify the spleen and stomach, and tonify qi and warm yang.<br>Cold-dampness in *Yangming*: This type of *Yangming* disease is mostly caused by damp pathogenic factors invading the spleen and stomach, and patients may experience chest and abdominal distension, sticky and greasy stools, and a thick and greasy tongue coating. Treatment can be carried out with cold medicines that warm yang and transform dampness, such as *Chenpi*, *Baizhu*, and *Sharen*, to warm and dry the spleen and stomach, and transform dampness and promote the movement of stagnation.<br>It should be noted that the use of cold medicines should be determined based on the specific symptoms, tongue and pulse conditions, and syndrome differentiation of the patients. At the same time, it is necessary to combine the compatibility and preparation of other medicines to achieve the effect of syndrome differentiation and treatment. When using cold medicines, the dosage and administration should be carefully controlled to avoid discomfort or adverse reactions. Therefore, it is recommended to consult a traditional Chinese medicine physician for more personalized and professional treatment advice when seeking medical treatment.(0 points) |

TORSS provides a correct explanation of the 'four methods of *Yangming*', with only one mistake in linking the 'acid-cold' method with the condition of 'li jie li zao'(Internal stagnation and dryness) in *Yangming*. There are also some irrelevant and extended contents. However, ChatGPT not only gives incorrect answers but also includes contraindications for the treatment of *Yangming* disease in its response.

问题 2：如何区分远血和近血，简述病机特点及治疗。（共 5 分）
Question 2: How to differentiate between distant bleeding and recent bleeding, and describe the characteristics and treatment of their pathogenesis. (5 points in total)

参考答案：《金匮》原文：'下血，先便后血，此远血也'；'下血，先血后便，此近血也'。因此，凡属脾气虚寒，气不摄血，症状为先便后血，血色黯紫，腹痛便溏，面色无华，神疲倦怠，四肢不温，舌淡脉细之远血者，治以黄土汤温脾摄血。方中灶中黄土温中涩肠止血；甘草、白术、附子温中健脾；干地黄、阿胶滋阴养血止血；黄芩苦寒反佐，防白术、附子温燥动血之弊。（3 分）属大肠湿热，迫血妄行，症见先血后便，下血鲜红或有粘液，大便不畅，苔黄腻，脉数之近血者，治以赤小豆当归散清热利湿，活血止血。方用赤小豆清热利湿，当归养血和血，待湿热去而下血自止。（2 分）

Answer: The original text from 'Jin Kui' states, 'In lower gastrointestinal bleeding, defecation precedes the presence of blood, indicating distant bleeding,' and 'In lower gastrointestinal bleeding, the presence of blood precedes defecation, indicating recent bleeding.' Therefore, for cases of spleen qi deficiency and cold, where the qi fails to contain the blood, the symptoms include defecation preceding the presence of blood, dark purple color of the blood, abdominal pain and loose stools, dull complexion, fatigue, cold extremities, pale tongue, and thin pulse. This is indicative of distant bleeding, and it should be treated with Huang Tu Tang to warm the spleen and contain the blood.

The formula contains ingredients such as Huang Tu to warm the intestines and stop bleeding, *Gan Cao*, *Bai Zhu*, and *Fu Zi* to warm the middle and strengthen the spleen, *Gan Di Huang* and *E Jiao* to nourish Yin and stop bleeding, and Huang Qin to counterbalance the cold properties and assist *Bai Zhu* and *Fu Zi* in warming and drying, to prevent the adverse effects of blood stagnation. (3 points)

For cases of large intestine damp-heat, which impels the blood to flow recklessly, the symptoms include the presence of blood preceding defecation, fresh red blood or mucus in the stool, constipation, yellow and greasy tongue coating, and rapid pulse. This indicates recent bleeding, and it should be treated with Chi Xiao Dou Dang Gui San to clear heat, promote diuresis, and activate blood circulation while stopping bleeding. The formula uses *Chi Xiao Dou* to clear heat and promote diuresis, and *Dang Gui* to nourish and activate blood. It should be used until the damp-heat is resolved and the bleeding stops naturally. (2 points)

Table 3 TORSS answers to question 2

| TORSS | TORSS |
|---|---|
| 远血和近血的主要区别在于出血的部位和病机特点。远血一般指消化道远端出血，如胃溃疡、十二指肠溃疡等，由于出血部位较远，出血速度较慢，因此常常出现黑色或暗红色大便。远血病机多为脾胃虚寒、气滞血瘀等，治疗时应注意温中散寒、活血化瘀。常用的方剂有归脾汤、补中益气汤等。近血一般指消化道近端出血，如食管炎、胃炎等，由于出血部位较近，出血速度较快，因此常常出现鲜红色大便。近血病机多为肝胃不和、湿热蕴结等，治疗时应注意疏肝和胃、清热利湿。常用的方剂有龙胆泻肝汤、左金丸等。 总之，远血和近血的病机特点和治疗方法有所不同，应根据出血部位和症状进行辨证施治。（3 分） | The main differences between upper gastrointestinal bleeding and lower gastrointestinal bleeding lie in the site of bleeding and the pathogenesis. Upper gastrointestinal bleeding generally refers to bleeding in the distal part of the digestive tract, such as gastric ulcers and duodenal ulcers. Due to the location of the bleeding being further away, the bleeding tends to be slower, resulting in the presence of black or dark red stools. The pathogenesis of upper gastrointestinal bleeding is often related to spleen and stomach deficiency-cold, Qi stagnation and blood stasis. Treatment should focus on warming the middle and dispersing cold, as well as promoting blood circulation and resolving stasis. Commonly used prescriptions include Guipi Decoction and Buzhong Yiqi Decoction. On the other hand, lower gastrointestinal bleeding generally refers to bleeding in the proximal part of |
| | the digestive tract, such as esophagitis and gastritis. Due to the location of the bleeding being closer, the bleeding tends to be faster, resulting in the presence of bright red stools. The pathogenesis of lower gastrointestinal bleeding is often related to disharmony between the liver and stomach, as well as damp-heat accumulation. Treatment should focus on soothing the liver and stomach, as well as clearing heat and promoting diuresis. Commonly used prescriptions include Longdan Xiegan Decoction and Zuo Jin Pill. In conclusion, upper gastrointestinal bleeding and lower gastrointestinal bleeding have different pathogenic characteristics and treatment methods. It is important to differentiate and treat based on the site of bleeding and symptoms.(3 points) |

Table 4 ChatGPT answers to question 2

| ChatGPT | ChatGPT |
|---|---|
| 远血和近血都是中医学中的病名，都属于痹证范畴。远血是指瘀血在远离病灶的部位引起的痛证，如瘀血阻滞经络，引起远端肢体麻木、疼痛。近血则是指瘀血在病灶附近引起的痛证，如创伤后局部淤血引起的疼痛。下面分别从病机特点和治疗方面介绍远血和近血。远血的病机特点是瘀血阻滞经络，气血不畅。临床表现为肢体局部麻木、疼痛，甚至肌肉萎缩、关节强直等。治疗上应当活血化瘀，通经活络。常用的中药有川芎、桃仁、红花等，常用的方剂有活血化瘀汤、四物汤等。另外，对于病情较重者，可选用刺络放血或拔罐等外治法。近血的病机特点是瘀血停滞在伤口周围，导致气血不畅，阻碍伤口愈合。临床表现为伤口肿胀、疼痛、发红等。治疗上应当活血化瘀，消肿止痛。常用的中药有三七、当归、蒲公英等，常用的方剂有活血化瘀汤、跌打丸等。另外，对于伤口较深或病情较重者，应当及时就医，进行手术处理或者局部抗生素药物治疗。 总之，远血和近血都是中医学中的病名，其治疗都应当以活血化瘀为主，同时结合具体病情进行个体化治疗。（0 分） | Distant blood and near blood are both disease names in traditional Chinese medicine and belong to the category of obstruction syndrome. Distant blood refers to the pain caused by blood stasis in a location far away from the lesion, such as blood stasis obstructing the meridians and causing numbness and pain in the distal limbs. Near blood, on the other hand, refers to the pain caused by blood stasis near the lesion, such as pain caused by local congestion after trauma. The characteristics of the pathogenesis and treatment of distant blood and near blood will be introduced separately below. The pathogenesis of distant blood is characterized by blood stasis obstructing the meridians and stagnant qi and blood. Clinically, it is manifested as local numbness and pain in the limbs, and even muscle atrophy and joint stiffness. The treatment should focus on promoting blood circulation and removing blood stasis, and commonly used Chinese herbs include Chuanxiong, Taoren, and Honghua. Commonly used formulas include Huoxue Huayu Tang and Siwu Tang. In addition, for patients with severe conditions, external treatments such as bloodletting or cupping can be used. The |
| | pathogenesis of near blood is characterized by blood stasis stagnating around the wound, leading to poor circulation of qi and blood and hindering wound healing. Clinical manifestations include swelling, pain, and redness around the wound. The treatment should focus on promoting blood circulation and removing blood stasis, as well as reducing swelling and relieving pain. Commonly used Chinese herbs include *Sanqi*, *Danggui*, and *Pugongying*, and commonly used formulas include Huoxue Huayu Tang and Dieda Wan. In addition, for patients with deep wounds or severe conditions, timely medical intervention such as surgical treatment or local antibiotic therapy is necessary. In conclusion, distant blood and near blood are both disease names in traditional Chinese medicine, and the treatment should focus on promoting blood circulation and removing blood stasis, while taking into account the specific condition for individualized treatment.(0 points) |

The symptom differentiation points and some pathogenesis in the answer of TORSS are correct, but the related herbal formulas are incorrect. ChatGPT's answers are incorrect.

**Example of Recall**

问题：请解释广义痰饮治疗原则："温药和之"的含义（5 分）
Question 1: Please explain the treatment principle of phlegm and retained fluid diseases: 'harmonizing with the warm medicine'.

参考答案：温药能振奋阳气，开发腠理，通行水道；'和之'有两层含义：一是不可太过温燥，二是勿专于温补。即使用温药时，应视病情恰当配合行、消、开、导、清之品。
Reference answer: Warm medicine can motivate Yang qi, induce perspiration, diuresis; *Harmonizing* has two meanings: one is not too warm, and the other is not dedicated to warming. That is, when using warm medicine, it should be appropriate to cooperate with the medicine of promoting circulation, dispersing, diaphoresis, purgation and clearing according to the condition.

TORSS 答案：'温药和之'这一治疗原则是在治疗广义痰饮病时的基本原则，它的含义主要是指使用温性的药物来调和病变。这里的'和'意味着调和、调节，而不是简单的增强或者抑制。在广义痰饮病的形成过程中，主要是因为肺、脾、肾三脏的阳气虚弱，导致气化不利，水液停聚而成。由于饮为阴邪，遇寒则聚，遇阳则行，得温则化。因此，治疗痰饮需要借助于温性的药物以振奋阳气，开发腑理，通调水道。阳气振奋，既可以温化饮邪，又可以绝痰饮滋生之源。

此外，'温药和之'也强调了对病情的全面考虑和治疗，不仅要针对病因进行治疗，还要考虑到病人的整体状况，使治疗和人体的正常生理机能相协调，达到真正的治疗效果。这种治疗原则体现了中医治疗疾病的整体观念和辨证施治的原则，也是中医学的一种重要治疗思想。

TORSS answer: The therapeutic principle of ‘harmonizing with the warm medicine’ is a fundamental approach in the treatment of general phlegm and retained fluid diseases. Its essence lies in using medicinal herbs with warm property to harmonize pathological changes. In this context, ‘harmonize’ implies reconciling and regulation, rather than simple enhancement or suppression.

In the formation process of generalized phlegm-damp diseases, the main cause is the weakness of the yang qi in the lungs, spleen, and kidneys, leading to unfavorable qi transformation and the accumulation of fluids. Since retained fluid is characterized as Yin pathogenic factors, it tends to aggregate in the presence of cold and disperse in the presence of warmth, being resolved effectively by warmth. Therefore, the treatment of retained phlegm and fluid involves the use of warm-property herbs to invigorate yang qi, open up sweat pore, and regulate water channels. Invigorating yang qi not only warms and transforms the pathogenic dampness but also eliminates the source of phlegm and retained fluid.

Furthermore, ‘harmonizing with the warm medicine’ emphasizes a comprehensive consideration and treatment of the patient's condition. The focus is not only on addressing the root cause of the disease but also taking into account the overall well-being of the individual. This approach aims to harmonize the treatment with the body's normal physiological functions, achieving genuine therapeutic effects. This therapeutic principle embodies the holistic perspective and the principles of syndrome differentiation and treatment in TCM, representing a significant therapeutic ideology in Chinese medical practice.

SPO-T RAG model recalled 15 SPO-Ts, in which 4 SPO-Ts (6,7,8,11) were closely related to the correct answer and selected by the self-reflection framework（bold font in the table）. The accurate recall helps the model got the right answer.

SOP-T 列表及召回评分（准确为 1，不准确为 0）:

SOP-T list and recall score (1 for accurate, 0 for inaccurate)

| | 召回的 SPO-T | Recalled SPO-T | score |
|---|---|---|---|
| 1 | Prescription - Warming Interior Formula - Huangqi Guizhi Wuwu Decoction - This paper introduces the therapeutic mechanism of Huangqi Guizhi Wuwu Decoction. This formula is applicable to deficiency of both the nourishing and defensive aspects of qi and blood, where the pathogenic factors of wind-cold invade and assail the blood vessels in the presence of weakness, leading to insufficient nourishment of the muscles and skin, manifesting symptoms such as numbness, mild aversion to cold, and a light red tongue. In contrast to wind-cold arthralgia, this condition, though resembling it, lacks a sense of pain. The pulse is slightly taut and tight, indicative of the characteristic of pathogenic stagnation in the blood vessels with congealed resistance. According to the 'Su Wen · Counterflow Theory,' the treatment for this condition should involve supplementing qi, warming the meridians, and promoting blood circulation to alleviate arthralgia. Therefore, Huangqi Guizhi Wuwu Decoction can treat this condition by tonifying qi and blood, and warming the meridians to facilitate unobstructed circulation. | 方剂-温里剂-温经散寒-黄芪桂枝五物汤-本文介绍了黄芪桂枝五物汤的证治机理。该方剂适用于营卫气血不足，风寒之邪乘虚客于血脉，导致肌肤失养，出现麻木、微恶风寒、舌淡等症状。与风痹不同的是，此证虽然类似风痹，但没有疼痛感，脉象微涩兼紧，表现为邪滞血脉、凝涩不通的特征。根据《素问·逆调论》，治疗该证应当益气温经，和血通痹。因此，黄芪桂枝五物汤可以通过补益气血、温经通络的作用来治疗该证。 | 0 |
| 2 | Seasonal Febrile Disease - accompanied symptoms and signs - phlegm and restrain fluid - clearing heat, removing phlegm, and resolving congestion- the following passage describes the application of the method of clearing heat, removing phlegm, and resolving congestion in various disease locations. For individuals with phlegm-heat obstructing the lungs, treatment may involve the use of medicines such as melon wilt and tian nan xing. For those with phlegm-heat causing chest congestion, medications like Xiao Xian Xiong Tang could be employed. In cases where phlegm-heat obstructs the orifices, additional herbs such as gallbladder nan xing and da huang may be added to formulas targeting heart-clearing and orifice-opening. Similarly, for individuals with phlegm-heat obstructing the liver meridian, medicines like niu huang and da huang may be included in formulas addressing heat-clearing and wind-dispelling. | 温病-兼夹证-痰饮-清热化痰开结-本节介绍了清热化痰开结法在不同病位的应用。对于痰热壅肺者，可以使用瓜蒌、川贝等药物进行治疗；对于痰热结胸者，可以使用小陷胸汤等药物进行治疗；对于痰热闭窍者，可以在清心开窍剂中加入胆南星、大黄等药物；对于痰热阻于肝经者，可以在清热息风剂中加入牛黄、大黄等药物。 | 1 |
| 3 | Prescription - Warming Interior Formula – Danggui Si Ni Decoction - the paper explores the therapeutic mechanisms of Danggui Si Ni Decoction in treating conditions such as blood deficiency, cold congealing in the meridians, and impaired blood circulation. The paper emphasizes that the primary causes of this syndrome are internal deficiency of yin-blood and external deficiency of yang-qi. The article also highlights the distinctions between this syndrome and conditions involving the depletion of the heart-kidney yang in the Shaoyin meridian and the extremities' rebellious cold due to internal accumulation of yin-cold. Finally, the paper underscores the treatment approach of warming the meridians, dispelling cold, and nourishing blood to promote meridian circulation. | 方剂-温里剂-温经散寒-当归四逆汤-本文主要讨论了当归四逆汤在治疗营血虚弱、寒凝经脉、血行不利等方面的证治机理。通过许宏的观点，指出阴血内虚和阳气外虚是导致该证的主要原因。文章还提到了该证与少阴心肾阳衰、阴寒内盛之四肢厥逆的不同之处。最后，强调了温经散寒、养血通脉的治疗方法。 | 0 |
| 4 | Prescription - Warming Interior Formula – Warming Liver Decoction - The main causes of this syndrome are insufficient liver and kidney function, the presence of cold pathogens in the liver meridian, and stagnation of Qi circulation. Cold, being a Yin pathogen, possesses characteristics of congealing and stagnation. When the functions of the liver and kidneys are weakened, cold pathogens easily invade the liver meridian, leading to impaired circulation of liver meridian Qi and the manifestation of symptoms such as testicular cold pain and lower abdominal discomfort. Treatment of this syndrome should focus on nourishing liver and kidney functions, dispersing cold stagnation, and promoting smooth Qi circulation. | 方剂-温里剂-温经散寒-暖肝煎-肝肾不足、寒客肝脉、气机郁滞是导致本证的主要原因。寒为阴邪，具有凝滞的特性，当肝肾功能不足时，寒邪容易侵袭肝脉，导致肝脉气机不畅，出现睾丸冷痛、小腹疼痛等症状。治疗该证应当补充肝肾功能，散寒凝，促进气机畅通。 | 0 |
| 5 | Prescription - Phlegm-expelling Decoction - Warming and Removing Cold Phlegm - San Zi Yang Qin Decoction-This formula is suitable for the elderly with constitutional weakness, spleen deficiency leading to stagnation of dampness from poor digestive function. The accumulated dampness transforms into phlegm, causing obstruction and stagnation of qi, disrupting the ascending and descending functions of the lungs. Symptoms such as coughing and chest tightness may manifest. Concurrently, phlegm-dampness affects the spleen, leading to spleen deficiency and poor digestion, resulting in reduced appetite, a white and greasy tongue coating, and a slippery pulse—typical manifestations of turbid phlegm. The therapeutic principles of this formula involve warming the lungs, transforming phlegm, descending qi, and aiding digestion. | 方剂-祛痰剂-温化寒痰-三子养亲汤-本方适用于年老体虚，脾运不佳导致停食生湿，湿聚成痰，痰浊阻滞，气机壅塞，肺失宣降，出现咳嗽、胸闷等症状；同时痰湿困脾，脾失健运，导致食少难消，舌苔白腻，脉滑，这些都是痰浊的表现。此方的治疗原则是温肺化痰，降气消食。 | 1 |
| **6** | **Golden Chamber Synopsis – phlegm and retain fluid – treatment principle – The treatment principles of phlegm and retain fluid diseases involves the use of warm-property herbal medicines.** | **金匮要略-痰饮-治疗原则-痰饮咳嗽病的治疗原则是使用温药** | **1** |
| **7** | **Internal medicine of TCM - phlegm and retain fluid diseases - The therapeutic principles for treating phlegm and retained fluid diseases primarily involve the use of warming and transforming methods. Through warming the yang to transform qi, the goal is to facilitate the easy transformation of dampness and the smooth flow of water, allowing the dispersion of phlegm with water. Specific treatment methods vary based on the differentiation of exterior, interior, deficiency, and excess conditions, employing warming, clearing, dispelling, and tonifying techniques accordingly. In cases of abundant phlegm-dampness, the focus is on eliminating the phlegm to address the symptoms; for conditions manifesting on the surface, warming and dispersing with sweating are recommended; for interior conditions, warming and transforming to benefit water metabolism are applied; deficiency conditions require tonification; and excess conditions call for purging methods. In cases involving a combination of excess and deficiency, a comprehensive approach of both dispelling and tonifying is employed. Moreover, if heat is mixed with phlegm, a combination of warming and clearing methods is necessary. In summary, the treatment of phlegm and retained fluid diseases demands flexible therapeutic strategies tailored to the specific condition of the illness.** | 中医内科学-痰饮-痰饮的治疗原则是以温化为主，通过温阳以化气，使饮易化且水易行，饮随水散。具体治法根据表虚实不同，采用温、清、消、补之法。对于水饮壅盛者，应祛饮以治标；邪在表者，应温散发汗；在里者，应温化利水；正虚者则进行补益；邪实者则采用攻破的方法。对于邪实正虚的情况，应综合消补。同时，如果饮热相杂，则需要温清并用。总之，痰饮的治疗需要根据病情采取灵活的治疗方法。 | **1** |
| **8** | **Golden Chamber Synopsis – phlegm and retain fluid – The principles for treating phlegm and retained fluid cough are mentioned in the context of the *Golden Chamber Synopsis*. Among them, the principle of 'harmonizing with warm medical herbs' is considered a fundamental approach for treating phlegm and retained fluid diseases. Clinical treatment requires specific analysis based on the pathological features. If it is a deficiency syndrome, it is necessary to differentiate between treating the spleen and treating the kidneys. If it is an excess syndrome, therapeutic methods such as promoting Xing, Xiao, Kai, Dao can be employed. '行' (Xing) refers to promoting the flow of qi, '消' (Xiao) means eliminating phlegm, '开' (Kai) indicates opening the yang qi, and '导' (Dao) signifies expelling phlegm through urination and defecation. In summary, when treating phlegm and retained fluid diseases, consideration should be given to the use of warming and tonifying herbs, as well as herbs with actions such as promoting circulation and eliminating dampness.** | 金匮要略-痰饮-金匮要略中提到了治疗痰饮咳嗽病的原则。其中，'温药和之'被认为是治疗痰饮病的基本原则。根据病理特征，临床治疗时需要具体分析。如果是正虚，需要区分治脾和治肾的方法；如果是邪实，则可以采用行、消、并、导等治法。行指行气，消指消除痰液，开指开阳气，导指通过大小便排出痰邪。总之，治疗痰饮咳嗽病时需要考虑温补和行消等药物的使用。 | **1** |
| 9 | Golden Chamber Synopsis - phlegm and retain fluid - The main content of this paper focuses on the etiology, pulse manifestations, and treatment of phlegm-damp cough. The article extensively discusses both the general and specific aspects of phlegm-damp, along with the treatment principles and specific therapeutic methods for phlegm-damp diseases. Phlegm-damp is a pathological condition caused by the invasion of dampness evils, with particular emphasis on the involvement of phlegm accumulation. There are four types of phlegm-damp, namely phlegm-damp, hanging-damp, overflowing-damp, and focal-damp. However, this paper primarily addresses the pathological changes of dampness evils lingering in the gastrointestinal tract. The paper proposes the treatment principle of using warming herbs to address phlegm-damp diseases, and introduces specific treatment methods such as warming, inducing sweating, promoting bowel movements, and draining downward, along with adaptive strategies based on symptomatic changes. These insights have significant guiding implications for subsequent clinical practice in the field, especially at the intersection of traditional Chinese medicine and artificial intelligence. | 金匮要略-痰饮-本篇正文是关于痰饮咳嗽病脉证并治的内容。文章重点讨论了痰饮的广义与狭义，以及痰饮病的治疗原则和具体治疗方法。痰饮是由饮邪引起的一种病症，其中支饮是其重点。痰饮包括痰饮、悬饮、溢饮和支饮四种类型，而本篇主要讨论的是饮邪停留在肠胃的病变。文章提出了以温药治疗痰饮病的治则，并介绍了温、汗、利、下等具体的治疗方法以及根据症状变化的应变法则。这对于后世临床实践具有重要的指导意义。 | 1 |
| 10 | Internal medicine of TCM – cold - In the treatment of the common cold in TCM, the guiding principle is to resolve from the surface, following the principle outlined in the 'Su Wen · Yin Yang Ying Xiang Da Lun,' which states, 'For those on the skin, induce sweating to expel it.' For wind-cold patterns, the treatment approach involves using pungent and warm methods to induce sweating. For wind-heat patterns, the treatment involves adopting pungent, cool, and resolving methods. In cases where dampness is mixed with pathogenic factors, the approach is to clear dampness and resolve the surface. In cases of deficiency-type cold, the strategy is to simultaneously resolve the surface and tonify the underlying deficiency. | 中医内科学-感冒-感冒的治疗原则是从表而解，遵循《素问·阴阳应象大论》中的'其在皮者，汗而发之'原则。对于风寒证，应采用辛温发汗的治疗方法；对于风热证，应采用辛凉清解的治疗方法；对于湿邪夹杂的情况，应清湿解表；对于虚体感冒，应同时解表和扶正。 | 0 |
| **11** | **Golden Chamber Synopsis – phlegm and retain fluid – This passage discusses the treatment principles for phlegm-dampness syndrome. The formation of phlegm-dampness syndrome is attributed to the weakness of the yang qi in the lungs, spleen, and kidneys, leading to the accumulation of water and fluids. The treatment of phlegm-dampness syndrome involves the use of warming herbs to invigorate yang qi, promote the smooth flow and regulation of water passages, and facilitate the elimination of pathogenic factors. Simultaneously, it is essential to be cautious in the use of warming herbs, avoiding excessive or overly drying and heating effects. The key to treating phlegm-dampness lies in harmonizing the body's yang qi, which is the fundamental approach to addressing the root cause.** | **金匮要略-痰饮-本条论述了治疗痰饮病的治疗原则。痰饮病的形成是由于肺脾肾三脏阳气虚弱，导致水液停聚。治疗痰饮病需要借助于温药来振奋阳气，开发通调水道，以疏通祛邪。同时，治疗痰饮需要注意温药的使用不可过度，也不可过于燥热。治疗痰饮的关键是调和人体阳气，这是治本之法。** | **1** |
| 12 | Internal medicine of TCM - phlegm and retain fluid diseases - Phlegm-dampness refers to the abnormal distribution and metabolism of body fluids, resulting in the main symptoms of accumulation in certain areas. It encompasses four categories: phlegm-dampness, suspended dampness, overflowing dampness, and obstructive dampness. Clinical manifestations of phlegm-dampness are diverse, including chronic bronchitis, bronchial asthma, exudative pleurisy, chronic gastritis, heart failure, and edema associated with nephritis. The formation and development of the theory of phlegm-dampness have laid a theoretical foundation. Zhang Zhongjing, in his work 'Jin Kui Yao Lue,' was the first to classify phlegm-dampness broadly into four categories and proposed corresponding treatment principles and prescriptions. From the Sui and Tang dynasties to the Jin and Yuan dynasties, a phlegm-dampness syndrome system gradually formed with phlegm at its core. After the Song dynasty, medical scholars conducted more in-depth research and summarization of the theory and treatment methods of phlegm-dampness. | 中医内科学-痰饮-痰饮是指体内水液输布、运化失常，停积于某些部位为主症的疾病。痰饮包括痰饮、悬饮、溢饮、支饮四类。痰饮的临床病种广泛，表现复杂，包括慢性支气管炎、支气管哮喘、渗出性胸膜炎、慢性胃炎、心力衰竭、肾炎水肿等。痰饮学说的形成和发展奠定了理论基础。张仲景在《金匮要略》中首次将广义痰饮分为四类，并提出了相应的治疗原则和方剂。隋唐至金元时期逐渐形成了以广义的痰为核心的痰饮病证体系。宋代以后，医家对痰饮的理论和治疗方法有了更深入的研究和总结。 | 1 |
| 13 | Internal medicine of TCM - phlegm and retain fluid diseases - This paper introduces and discusses the treatment principles and essence of phlegm-dampness syndrome based on two classical Chinese medical texts, *Synopsis of Golden Chamber* and *Water, Dampness, Phlegm, Fluids*. In *Synopsis of Golden Chamber,* the primary approach to treating phlegm-dampness syndrome is through the use of warm-tonifying herbs. On the other hand, *Water, Dampness, Phlegm, Fluids* analyzes the essence and treatment principles of phlegm-dampness. Additionally, the literature *On the essence of phlegm and drink* provides further insights into the essence of phlegm-dampness. Through the introduction and analysis of these texts, a better understanding and treatment approach for phlegm-dampness syndrome can be achieved in the intersection of traditional Chinese medicine and artificial intelligence. | 中医内科学-痰饮-本文介绍了《金匮要略方论》和《水、湿、痰、饮》两篇文献对于痰饮证的治疗原则和实质的探讨。其中，《金匮要略方论》认为痰饮证应该以温药治疗为主，《水、湿、痰、饮》则对痰饮的实质和治疗原则进行了分析。另外，还有一篇文献《论痰饮实质》对痰饮的实质进行了论述。通过这些文献的推介和分析，可以更好地理解和治疗痰饮证。 | 1 |
| 14 | *Treatise of Febrile Diseases* -The name '*Treatise of Febrile Diseases*' is derived from the term 'Shanghan,' and its meaning has both broad and narrow interpretations. In a broad sense, Shanghan encompasses all febrile diseases caused by external pathogens, while in a narrow sense, it specifically refers to diseases arising from the invasion of wind-cold factors. The text primarily discusses Shanghan, Zhongfeng, Wenbing(seasonal febrile diseases), and others, falling within the category of broad Shanghan. It is crucial to emphasize that the concept of 'Shanghan' as discussed in ' *Treatise of Febrile Diseases*' is entirely different from the Western medical understanding of 'typhoid fever.'' | 伤寒论-《伤寒论》以伤寒命名，其含义广义和狭义有所区分。广义伤寒是指一切外感热病的总称，而狭义伤寒则是指外感风寒，感而即发的疾病。书中主要讨论了伤寒、中风、温病等，属于广义伤寒的范畴。需要强调的是，《伤寒论》所论的伤寒病与西医学中的'伤寒'含义完全不同。 | 0 |
| 15 | Golden Chamber Synopsis – phlegm and retain fluid – This paper primarily discusses diseases related to phlegm-fluid retention and cough, emphasizing the significance of phlegm-fluid retention. The formation of phlegm-fluid retention is associated with the spleen, lungs, and kidneys, with a particularly close relationship to the spleen. Phlegm-fluid retention can be broadly and narrowly classified. Broadly, it includes four types: phlegm-fluid retention, suspended fluid, overflowing fluid, and branchial fluid. Narrowly, phlegm-fluid retention specifically refers to lesions in the gastrointestinal tract.<br>The affected areas of phlegm-fluid retention are mainly in the gastrointestinal tract, subcostal region, thoracic diaphragm, and the interstitial tissues of the limbs, and it can also impact the five viscera. The classification of the four types of fluids is based on the location of water retention and the main symptoms. Different prescriptions can be used interchangeably. The principle of treating phlegm-fluid retention is primarily based on the use of warming drugs, and the treatment methods should also be adapted according to conditions such as deficiency or excess, cold or heat. | 金匮要略-痰饮-本篇主要论述痰饮和咳嗽的疾病，并强调了痰饮的重要性。痰饮的形成与脾、肺、肾有关，尤其是与脾的关系密切。痰饮可以广义和狭义来分类，广义痰饮包括痰饮、悬饮、溢饮和支饮四种，而狭义痰饮仅指胃肠的病变。痰饮的病变部位主要在胃肠、胁下、胸隔和肢体肌肤之间，并且可以影响五脏。四饮是根据水饮停聚的部位和主要症状进行分类的，同时不同的方剂可以互相通用。治疗痰饮的原则是以温药和之为主，同时还需根据虚实、寒热等情况采用相应的治疗方法。对于支饮服用小青龙汤后的变证处理，是辨证论治的示范。本篇中涉及的方剂在其他篇和《伤寒论》中也有出现，学习时需要相互对勘，加深理解，以便灵活应用。 | 1 |